%% file: main.tex
\documentclass{article}

\usepackage{microtype}
\usepackage{graphicx}
\usepackage{subfigure}
\usepackage{booktabs} %

\usepackage{hyperref}

\usepackage[accepted]{icml2021}

\icmltitlerunning{Clusterability as an Alternative to Anchor Points}

\input{math}

\usepackage{listings}

\usepackage{graphicx,subfigure,epsfig}
\usepackage{comment}
\usepackage{xcolor,enumerate}

\usepackage{threeparttable}

\usepackage{multirow}

\usepackage{tcolorbox}
\definecolor{grey}{rgb}{0.33, 0.33, 0.33}

\usepackage{algorithmic}
\usepackage{algorithm}
\usepackage{enumitem}

\ifodd 0
\newcommand{\rev}[1]{{\color{blue}#1}}

\newcommand{\clar}[1]{\textbf{\color{green}(NEED CLARIFICATION: #1)}}

\else
\newcommand{\rev}[1]{#1}
\newcommand{\com}[1]{}
\newcommand{\clar}[1]{}
\newcommand{\response}[1]{}
\fi

\newcommand{\algcom}[1]{\textsl{\color{blue}{\footnotesize #1}}}

\newtheorem{thm}{Theorem}

\newtheorem{lem}{Lemma}
\newtheorem{cor}{Corollary}

\newtheorem{defn}{Definition}
\newtheorem{rmk}{Remark}

\newtheorem{ap}{Assumption}

\newcommand{\ours}{\textsf{HOC}}

\usepackage{lipsum}

\begin{document}

\twocolumn[
\icmltitle{Clusterability as an Alternative to Anchor Points\\ When Learning with Noisy Labels}

\begin{icmlauthorlist}
\icmlauthor{Zhaowei Zhu}{UCSC}
\icmlauthor{Yiwen Song}{BUPT}
\icmlauthor{Yang Liu}{UCSC}
\end{icmlauthorlist}

\icmlaffiliation{UCSC}{Department of Computer Science and Engineering, University of California, Santa Cruz, CA, USA}
\icmlaffiliation{BUPT}{Beijing University of Posts and Telecommunications, Beijing, China}

\icmlcorrespondingauthor{Yang Liu}{yangliu@ucsc.edu}

\icmlkeywords{Clusterability, learning with label noise, noise transition matrix}

\vskip 0.3in
]

\printAffiliationsAndNotice{}  %

\input{src/abs}
\input{src/intro}

\input{src/pre}

\input{src/errEst}

\input{src/guarantee}

\input{src/exp}
\input{src/conclusion}

\textbf{Acknowledgment}~
This work is partially supported by the National Science Foundation (NSF) under grant IIS-2007951 and the Office of Naval Research under grant N00014-20-1-22.

\clearpage
\newpage
\bibliography{ref}
\bibliographystyle{icml2021}

\clearpage
\newpage
\appendix
\input{src/appendix}

\end{document}

%% file: math.tex
\usepackage{amsmath,amsfonts,bm,amsthm,dsfont}

\def\eqref#1{equation~\ref{#1}}

\def\1{\bm{1}}
\def\0{\bm{0}}

\def\vc{{\bm{c}}}

\def\ve{{\bm{e}}}
\def\vf{{\bm{f}}}

\def\vp{{\bm{p}}}

\def\vu{{\bm{u}}}

\def\vx{{\bm{x}}}
\def\vy{{\bm{y}}}

\def\mA{{\bm{A}}}
\def\mB{{\bm{B}}}
\def\mC{{\bm{C}}}
\def\mD{{\bm{D}}}

\def\mI{{\bm{I}}}

\def\mQ{{\bm{Q}}}

\def\mS{{\bm{S}}}
\def\mT{{\bm{T}}}
\def\mU{{\bm{U}}}

\def\vlmd{{\bm{\lambda}}}

\DeclareMathAlphabet{\mathsfit}{\encodingdefault}{\sfdefault}{m}{sl}
\SetMathAlphabet{\mathsfit}{bold}{\encodingdefault}{\sfdefault}{bx}{n}

\DeclareMathOperator*{\argmax}{arg\,max}
\DeclareMathOperator*{\argmin}{arg\,min}

\newcommand{\PP}{\mathbb P}

\newcommand{\BR}{\mathds 1}

\usepackage{tcolorbox}
\definecolor{grey}{rgb}{0.33, 0.33, 0.33}

\newcommand{\SPL}{CORES$^2$}

\newcommand{\squishlist}{
\begin{list}{{{\small{$\bullet$}}}}
{\setlength{\itemsep}{1pt}      \setlength{\parsep}{0pt}
\setlength{\topsep}{-2pt}       \setlength{\partopsep}{0pt}
\setlength{\leftmargin}{1em} \setlength{\labelwidth}{1em}
\setlength{\labelsep}{0.5em} } }
\newcommand{\squishend}{  \end{list}  }

\makeatletter
\renewcommand*\env@matrix[1][*\c@MaxMatrixCols c]{%
  \hskip -\arraycolsep
  \let\@ifnextchar\new@ifnextchar
  \array{#1}}
\makeatother

%% file: src/abs.tex
\begin{abstract}
The label noise transition matrix, characterizing the probabilities of a training instance being wrongly annotated, is crucial to designing popular solutions to learning with noisy labels. Existing works heavily rely on \rev{finding} ``anchor points'' or their approximates, defined as instances belonging to a particular class almost surely. Nonetheless, finding anchor points remains a non-trivial task, and the estimation accuracy is also often throttled by the number of available anchor points. In this paper, we propose an alternative option to the above task. Our main contribution is the discovery of an efficient estimation procedure based on a clusterability condition. We prove that with clusterable representations of features, using up to third-order consensuses of noisy labels among neighbor representations is sufficient to estimate a unique transition matrix. Compared with methods using anchor points, our approach uses substantially more instances and benefits from a much better sample complexity. We demonstrate the estimation accuracy and advantages of our estimates using both synthetic noisy labels (on CIFAR-10/100) and real human-level noisy labels (on Clothing1M and our self-collected human-annotated CIFAR-10).
\rev{Our code and human-level noisy CIFAR-10 labels are available at \url{https://github.com/UCSC-REAL/HOC}.}
\end{abstract}

%% file: src/intro.tex
\section{Introduction}

Training deep neural networks (DNNs) relies on the large-scale labeled datasets while they often include a non-negligible fraction of wrongly annotated instances.
\rev{The corrupted patterns tend to be memorized by the over-parameterized DNNs \cite{xia2021robust,han2020survey}, and lead to unexpected and disparate impacts \cite{liu2021importance}.} %

A variety of approaches were proposed to address the problem of learning with noisy labels. The implementations of a major line of them, e.g., \citet{patrini2017making,xiao2015learning,xia2020parts,berthon2020confidence,xia2019anchor,yao2020dual,li2021provably}, depend on accurate knowledge of the \emph{noise transition matrix} $\mT$, which characterizes the probabilities of a training example being wrongly annotated.
It has been show that \cite{liu2015classification,patrini2017making},  with perfect knowledge of $\mT$, the minimizer of a corrected or reweighted expected risk (loss) defined on the noisy distribution is the same as the minimizer of the true expected risk (loss) of the clean distribution. %
These results clearly established the power and benefits of knowing $\mT$.

Estimating $\mT$ is challenging without accessing clean labels. Existing works on estimating $\mT$ often rely on finding a number of high-quality anchor points \cite{scott2015rate,liu2015classification,patrini2017making}, or approximate anchor points \cite{xia2019anchor}, which are defined as the training examples that belong to a particular class almost surely.
To find the anchor point, a model needs to be trained to accurately characterize the noisy label distribution. This model will help inform the selection of anchor points. {Again relying on this model, $\mT$ is then estimated using posterior noisy label distributions of the anchor points.}

While the anchor point approach observes a significant amount of successes, it suffers from several limitations: 1) accurately fitting noisy distributions is challenging when the number of label classes is high; 2) the number of anchor points restricts the estimation accuracy; and 3) it lacks the flexibility to extend to more complicated noise settings. 
\rev{Other methods such as confident learning \cite{northcutt2017learning,northcutt2021confident} may not explicitly identify anchor points, but they still need to fit the noisy distributions and find some ``confident points'', thus suffer from the above limitations.}

In this paper, we provide an alternative to estimate $\mT$ without resolving to anchor points. %
The only requirement we need is clusterability, i.e., the \textbf{two} nearest-neighbor representations of a training example and the example itself belong to the same true label class.
Our main contributions summarize as follows:
\vspace{-2pt}
\squishlist
    \item Based on the clusterability condition, we propose a novel $\mT$ estimator by exploiting a set of high-order consensuses information among neighbor representations' noisy labels. Compared with the methods using anchor points, our estimator uses a much larger set of training examples and benefits from a much better sample complexity. 
    \item We prove that using up to third-order consensuses is sufficient to identify the true noise transition matrix uniquely. 
    \item Extensive empirical studies on CIFAR-10/100 datasets with synthetic noisy labels, the Clothing1M dataset with real-world human noise, and the CIFAR-10 dataset with our self-collected human annotations, demonstrate the advantage of our estimator. %
    \item Open-source contribution and flexible extension: we will contribute to the community 1) a generically applicable and light tool for fast estimation of the noise transition matrix. This flexible tool has the potential to be applied to more sophisticated noise settings, including instance-dependent ones (Section~\ref{sec:IDN}). 2) A noisy version of the CIFAR-10 dataset with human-level label noise.
\squishend

\subsection{Related Works}

In the literature of learning with label noise, a major set of works focus on designing \emph{risk-consistent} methods, i.e., performing empirical risk minimization (ERM) with specially designed loss functions on noisy distributions leads to the same minimizer as if performing ERM over the corresponding unobservable clean distribution.
The \emph{noise transition matrix} is a crucial component for implementing risk-consistent methods, e.g., loss correction \cite{patrini2017making}, loss reweighting \cite{liu2015classification}, label correction \cite{xiao2015learning} and unbiased loss \cite{natarajan2013learning}. To a certain degree, the knowledge of it also helps tune hyperparameters in other approaches, e.g., label smoothing \cite{lukasik2020does}.
As introduced previously, anchor points are critical for estimating the transition matrix in above mentioned existing methods - we further elaborate this in Section \ref{sec:lc}.

Some recently proposed risk-consistent approaches do not require the knowledge of transition matrix, including: $L_{\sf DMI}$ \cite{xu2019l_dmi} based on an information theoretical measure, peer loss \cite{liu2019peer} \rev{by punishing over-agreements with noisy labels, robust $f$-divergence \cite{wei2021when}}, and  CORES$^2$ \cite{sieve2020} built on a confidence-regularizer.
However, to principally handle a more complicated case when the noise transition matrix depends on each feature locally, i.e., instance-dependent noise, the ability to estimate \emph{local transition matrices} remains a significant and favorable property. %
Examples include the potential of applying local transition matrices to different groups of data \cite{xia2020parts}, using confidence scores to revise transition matrices \cite{berthon2020confidence}, and estimating the second-order information of local transition matrices \cite{zhu2020second}. Thus we need an estimation approach that scales and generalizes well to these situations.

As a growing literature, we are aware of other promising approaches that do not rely on the estimation of $\mT$, e.g., focusing on the numerical property of loss functions and designing bounded loss functions \cite{amid2019robust,amid2019two,zhang2018generalized,wang2019symmetric,gong2018decomposition,ghosh2017robust,shu2020learning}, using sample selection to pick up reliable instances from the dataset \cite{jiang2017mentornet,han2018co,yu2019does,yao2020searching,wei2020combating}, among many more. We compare to some of the popular ones using experiments.

%% file: src/pre.tex
\section{Preliminaries}\label{Sec:pre}

This section introduces the preliminaries, including problem formulation, anchor points, and the clusterability condition.

\subsection{Our Setup}
We summarize the important definitions as follows.

\textbf{Clean/Noisy distribution~}
The traditional classification problem with clean labels often builds on a set of $N$ training examples denoted by $D:=\{( x_n,y_n)\}_{n\in  [N]},$ where $[N] := \{1,2,\cdots,N\}$. Each example $(x_n,y_n)$ could be seen as a snapshot of random variable $(X,Y)$ drawn from a clean distribution $\mathcal D$. Let $\mathcal X$ and $\mathcal Y$ denote the space of feature $X$ and label $Y$, respectively. %
In our considered weakly-supervised classification problem, instead of having access to the clean dataset $D$, the learner could only obtain a noisy dataset $\widetilde D :=\{( x_n, \tilde y_n)\}_{n\in  [N]},$ where the noisy label $\tilde y_n$ may or may not be the same as $y_n$.
Noisy examples $(x_n,\tilde y_n)$ are generated according to random variables $(X,\widetilde Y)$ drawn from a distribution $\widetilde{\mathcal{D}}$.

\textbf{Noise transition matrix~}
We model the relationship between $(X,Y)$ and $(X,\widetilde Y)$ using a noise transition matrix $\mT(X)$, where each element $T_{ij}(X)$ represents the probability of mislabeling a clean label $Y=i$ to the noisy label $\widetilde Y = j$, i.e. $T_{ij}(X):=\mathbb P(\widetilde{Y}=j|Y=i,X).$
We call $\mT(X)$ the \emph{local} transition matrix in this paper since it is defined for a particular feature $X$.
Most of the literature would focus on the case where the noise is independent of feature $X$: $\mT(X) \equiv \mT$. 
The knowledge of $\mT$ enables a variety of learning with noisy label solutions. Below we illustrate solutions with the celebrated loss correction approach \cite{natarajan2013learning,patrini2017making}.

\textbf{The learning task~} The classification task aims to identify a classifier $\vf: \mathcal X \rightarrow \mathcal Y$ that maps $X$ to $Y$ accurately. 
We focus on minimizing the empirical risk using DNNs with respect to the cross-entropy (CE) loss defined as
$
    \ell(\vf(X),Y) = - \ln(f_X[Y]), ~ Y \in [K],
$
where $f_X[Y]$ denotes the $Y$-th component of column vector $\vf(X)$ and $K$ is the number of classes.

\subsection{Loss Correction and Estimating \texorpdfstring{$\mT$}{}}\label{sec:lc}

In the popular loss correction approach \cite{patrini2017making}, when the noise transition matrix is known, forward or backward loss correction can be applied to design a corrected loss. 
For example, the forward loss correction function can be designed as:
$
\ell^{\rightarrow}(\vf(X),\widetilde Y):= \ell(\mT^\top \vf(X),\widetilde Y),
$
where $\mT^\top$ denotes the transpose of matrix $\mT$.
If $\mT$ is perfectly known in advance, it can be shown that the minimizer of the corrected loss under the noisy distribution is the same as the minimizer of the original loss $\ell$ under the clean distribution \cite{patrini2017making}.

We would like to emphasize that in addition to loss correction, the knowledge of noise transition matrices is potentially useful in other approaches, especially when dealing with the challenging instance-dependent label noise where $\mT(X)$ differs for different $X$. For example, it was shown that knowing $\mT(X)$ helps improve the robustness of peer loss when the noise transition matrix differs across instances \cite{zhu2020second}, and can help improve fairness guarantees when label noise is group-dependent \cite{wang2020fair}. Knowing $\mT$ also tends to be helpful in setting hyperparameters in sample selection \cite{han2018co} and label smoothing \cite{lukasik2020does,wei2021understanding}.

\textbf{Estimating $\mT$ with anchor points~}
The traditional approach for estimating $\mT$ relies on anchor points \cite{scott2015rate,liu2015classification}, which are defined as the training examples ($X$s) that belong to a specific class almost surely.
Formally, an $x$ is an anchor point for the class $i$ if $\PP(Y = i|X=x)$ is equal to one or close to one \cite{xia2019anchor}. Further, if $\PP(Y = i|X = x)=1$, we have
$
\PP(\widetilde Y=j | X=x) = \sum_{k\in[K]} T_{kj} \PP(Y=k|X=x) = T_{ij}.
$
The matrix $\mT$ can be obtained via estimating the noisy class posterior probabilities for anchor points heuristically \cite{patrini2017making} or theoretically \cite{liu2015classification}. 

While the anchor point approach observes a significant amount of successes, this method suffers from three major limitations:
\squishlist
\item The implementation of it requires that the trained model can perfectly predict the probability of the noisy labels, which is challenging when the number of classes is high, and when the number of training instances is limited. 
\item The number of available and identifiable anchor points can become a bottleneck even if the posterior distribution can be perfectly learned.
\item The lack of flexibility to zoom into a subset of training data also limits its potential to be applied to estimate local transition matrices for more challenging instance-dependent settings \cite{xia2019anchor}.
\squishend

\subsection{Clusterability}

The alternative we are seeking builds on the notion of clusterability. Intuitively, clusterability implies that two instances are likely to have the same labels if they are close to each other \cite{gao2016resistance}. To facilitate the discovery of close-by instances, our solution will resolve to representation learning \cite{bengio2013representation}. 
Recent literature shows, even though label noise makes the model generalizes poorly, it still induces good representations \cite{li2020noisy}.
{Formally, for a neural network with both convolutional layers and linear layers, e.g., ResNet \cite{he2016deep}, we denote the convolution layers by function $\vf_{\sf conv}$ and the representations by $\bar X := \vf_{\sf conv}(X)$.}
With the above, we define $k$-Nearest-Neighbor ($k$-NN) label clusterability\footnote{Distances are measured between representations. Feature $x_n$ and its representation $\bar x_n$ refer to the same data point in different views.} as:
\begin{defn}[$k$-NN label clusterability]\label{def:cluster}
We call a dataset $D$ satisfies $k$-NN label clusterability if $\forall n \in [N]$, the representation $\bar x_n$ and its $k$-Nearest-Neighbor $\bar x_{n_1}, \cdots, \bar x_{n_k}$ belong to the same true class.
\end{defn}

\begin{figure}[!t]
    \centering
    \includegraphics[width=0.43\textwidth]{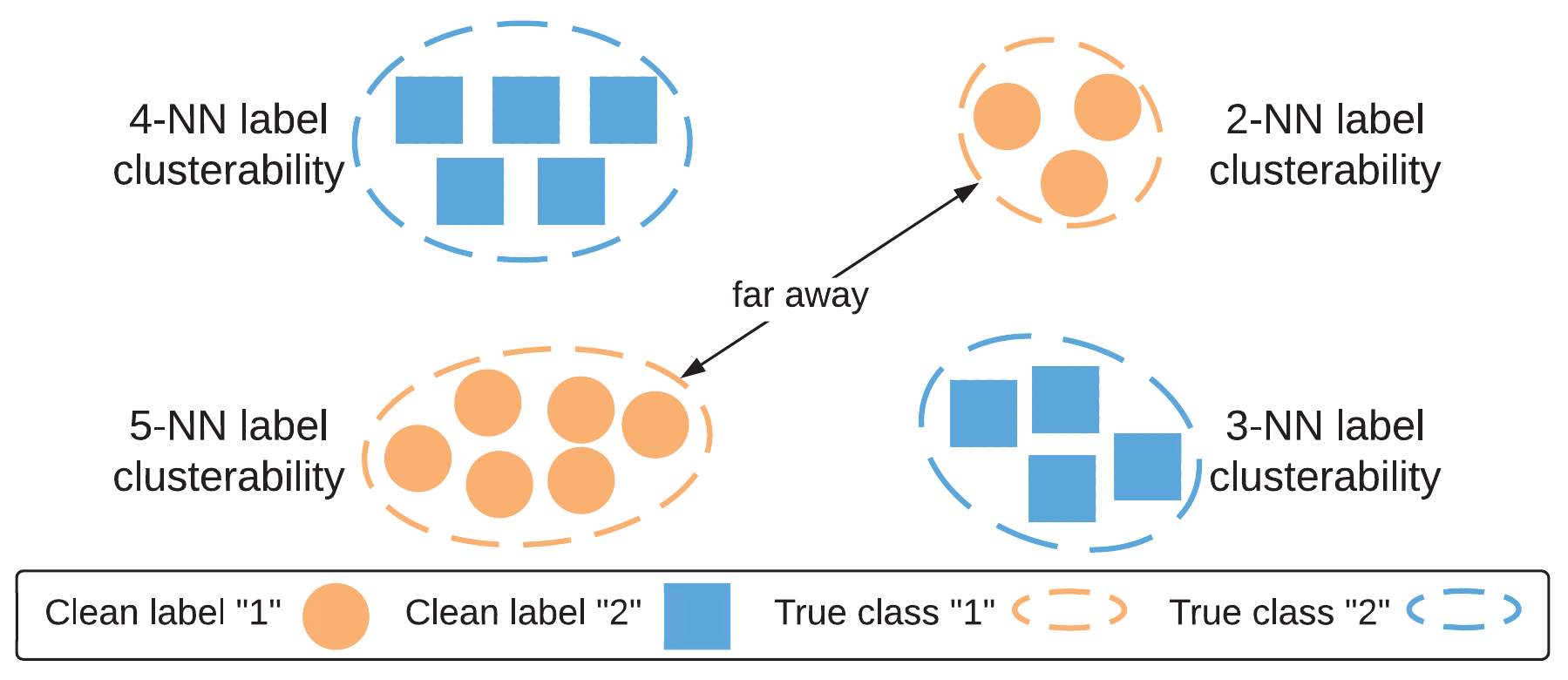}
    \vspace{-5pt}
    \caption{Illustration of $k$-NN label clusterability.  
    }
    \vspace{-7pt}
    \label{fig:illustCluster}
\end{figure}

\rev{
See Figure~\ref{fig:illustCluster} for an illustration of the $k$-NN clusterability. 
There are three primary properties of the definition:
\begin{itemize}[itemsep= 1 pt,topsep = -2 pt, leftmargin = 5mm]
    \item The $k_1$-NN label clusterability condition is harder to satisfy than $k_2$-NN label clusterability when $k_1 > k_2$;
    \item The cluster containing the same clean labels is not required to be a continuum, e.g., in Figure~\ref{fig:illustCluster}, two clusters of class ``1'' can be far away;
    \item Our $k$-NN label clusterability only requires the existence of these feasible points, i.e., specifying the true class is not necessary.
\end{itemize}
}

The $k$-NN label clusterability likely holds in many tasks, such as image classification when features are well-extracted by convolutional layers \cite{han2019deep,ji2019invariant,kolesnikov2019revisiting} \rev{and each feature belongs to a unique true class.
The high-level intuition is that similar representations should belong to the same label class. 
One can consider a label generation process \cite{feldman2020does,liu2021importance} where the feature distribution is modeled as a mixture of many disjoint sub-distributions, and the labeling function maps each sub-distribution to a unique label class. Therefore, samples from the same sub-distribution have the same true label.
In this paper, instead of requiring identical labels for a big cluster defined by a large $k$, we will only require the $\mathbf{2}$ nearest neighbors to have the same clean labels with the example itself, i.e., \emph{$2$-NN label clusterability}. Its feasibility will be demonstrated in Section~\ref{sec:feasibility}.

\textbf{Comparison to anchor points}~
The anchor point approach relies on training a classifier to identify anchor points and the corresponding true class. Our label clusterability definition \emph{does not require the knowledge of true label class} as claimed in the third property. Moreover, if good representations are available apriori, our method is \emph{model-free}.}

Next, we will elaborate our proposed $\mT$ estimator leveraging $2$-NN label clusterability. 
Relaxation of $2$-NN label clusterability is discussed in Appendix~\ref{supp:relax2nn}.

%% file: src/errEst.tex
\section{The Power of High-Order Consensuses} %

We now present our alternative to estimate $\mT$. Our idea builds around the concept of using high-order consensuses of the noisy labels $\widetilde Y$s among each training instance and its 2-NN. In this section, we consider the case when $\mT(X)$ is the same for different $X$, i.e., $\mT(X) \equiv \mT$.

\subsection{Warm-up: A Binary Example}

For a gentle start, consider binary cases ($K=2$) with classes $\{1,2\}$.
Short-hand error rates $e_1 := T_{12}:=\PP(\widetilde Y = 2|Y=1)$, $e_2:=T_{21}:=\PP(\widetilde Y = 1|Y=2)$. 
$p_1 := \PP(Y=1)$ denotes the clean prior probability of class-$1$.

We are inspired by the matching mechanism for binary error rates estimation \cite{liu2017machine,liu2020surrogate}.
Intuitively, with $1$-NN label clusterability, for two representations in the same dataset with minimal distance, their labels should be identical. Otherwise, we know there must be exactly one example with the corrupted label.
Similarly, if $k$-NN label clusterability holds, by comparing the noisy label of one representation with its $k$-NN, we can write down the probability of the $k+1$ noisy label consensuses (including agreements and disagreements) as a function of $e_1,e_2,p_1$.

\textbf{Going beyond votes from $k$-NN noisy labels}~
To infer whether the label of an instance is clean or corrupted, one could use the $2$-NN of this instance and take a majority vote. For example, if the considered instance has the label ``1'' and the other two neighbors have the label ``2'', it can be inferred that the label of the considered instance is corrupted since ``2'' is in the majority.
Nonetheless, this inference would be wrong when the $2$-NN are corrupted.
Increasing accuracy of the naive majority vote \cite{liu2015online} or other inference approaches \cite{liu2012variational} requires stronger clusterability that more neighbor representations should belong to the same clean class. Our approach goes beyond simply using the votes among $k$-NNs. Instead, we will rely on the statistics of high-order consensuses among the $k$-NN noisy labels. As a result, our method enjoys a robust implementation with only requiring $2$-NN label clusterability.

\textbf{Consensuses in binary cases}~
We now derive our approach for the binary case to deliver our main idea. We present the general form of our estimator in the next subsection.
Let $\widetilde Y_1$ be the noisy label of one particular instance, $\widetilde Y_2$ and $\widetilde Y_3$ be the noisy labels of its nearest neighbor and second nearest neighbor.
With $2$-NN label clusterability, their clean labels are identical, i.e. $Y_1 = Y_2 = Y_3$.
For $\widetilde Y_1$, noting
$
\PP(\widetilde Y_1 = j) = \sum_{i\in[K]}\PP(\widetilde Y_1 = j| Y_1 = i)\cdot \PP(Y_1=i),
$
we have the following \textbf{two} first-order equations:
\vspace{-3pt}
\begin{equation*}
    \begin{split}
        \PP(\widetilde Y_1=1) &= p_1(1-e_1) + (1-p_1)e_2, \\
        \PP(\widetilde Y_1=2) &= p_1 e_1 + (1-p_1)(1-e_2). 
    \end{split}
\end{equation*}
\vspace{-3pt}
For the second-order consensuses, we have

{\vspace{-17pt}\small
\begin{align*}
     &\PP(\widetilde Y_1 = j_1,\widetilde Y_2 = j_2) \\
    \overset{(a)}{=}&\sum_{i\in[K]}\PP(\widetilde Y_1 = j_1,\widetilde Y_2 = j_2| Y_1= i, Y_2 = i)\cdot \PP(Y_1=i) \\
    \overset{(b)}{=}&\sum_{i\in[K]}\PP(\widetilde Y_1 = j_1| Y_1 = i)\cdot\PP(\widetilde Y_2 = j_2| Y_2 = i)\cdot \PP(Y_1=i),
\end{align*}\vspace{-17pt}}

where equality $(a)$ holds due to the $2$-NN label clusterability, i.e., $Y_1 = Y_2 (=Y_3)$ w.p. 1, and equality $(b)$ holds due to the conditional independency between  $\widetilde Y_1$ and $\widetilde Y_2$ given their clean labels. In total, there are \textbf{four} second-order equations for different combinations of $\widetilde{Y}_1,\widetilde{Y}_2$, e.g., %
\vspace{-3pt}
\begin{equation*}
    \begin{split}
        \PP(\widetilde Y_1=1, \widetilde Y_2=1) &= p_1(1-e_1)^2 + (1-p_1)e_2^2, \\
        \PP(\widetilde Y_1=1, \widetilde Y_2=2) &= p_1(1-e_1)e_1 + (1-p_1)e_2(1-e_2).
    \end{split}
\end{equation*}
\vspace{-3pt}

\begin{figure}[!t]
    \centering
    \includegraphics[width=0.41\textwidth]{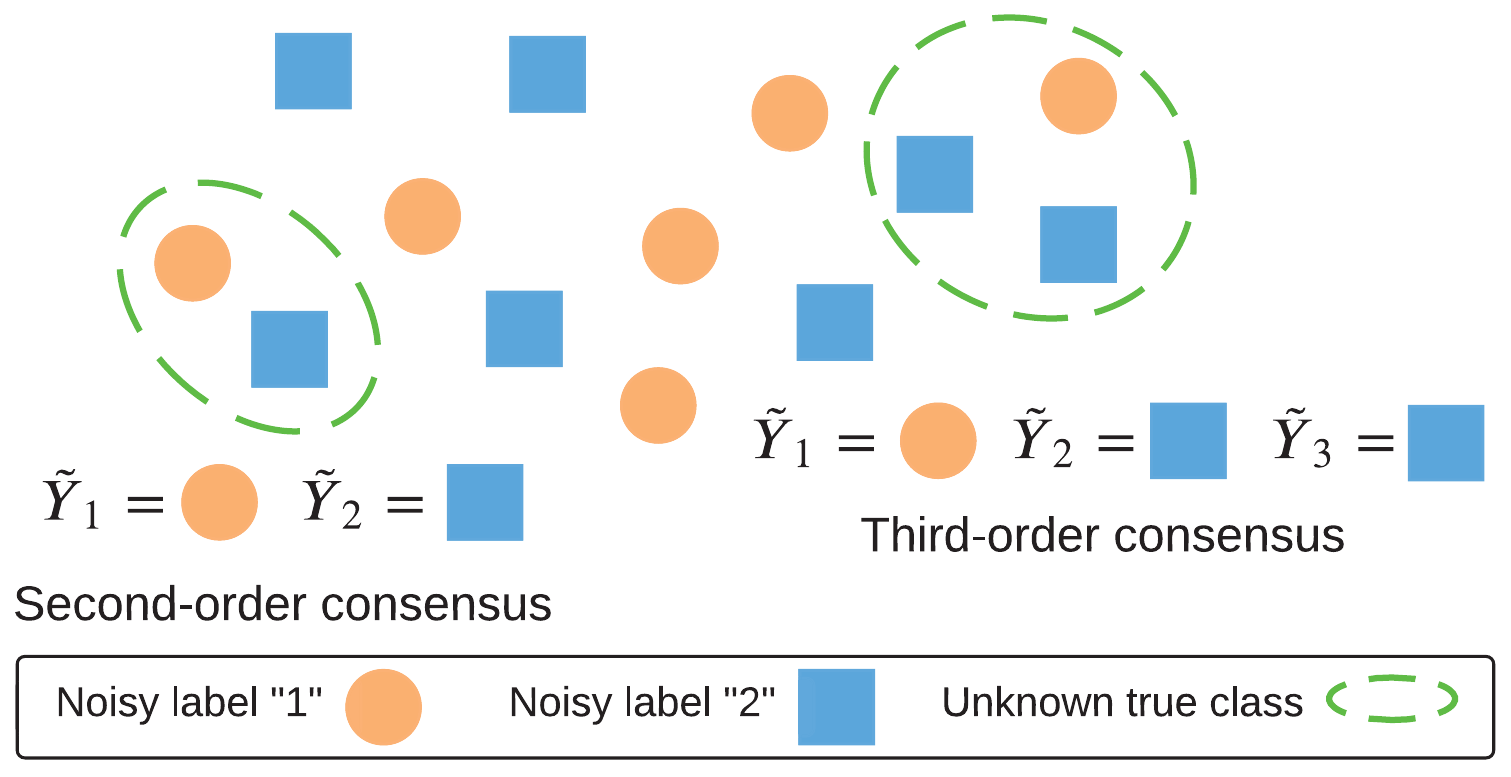}
    \vspace{-5pt}
    \caption{Illustration of high-order consensuses. 
    }%
    \vspace{-7pt}
    \label{fig:illustHOC}
\end{figure}

Similarly, given $Y_1 = Y_2 = Y_3$, there are \textbf{eight} third-order equations defined for consensuses among %
$\widetilde Y_1,\widetilde Y_2,\widetilde Y_3$ , e.g.,
\[
    \PP(\widetilde Y_1=1, \widetilde Y_2=1, \widetilde Y_3=1) = p_1(1-e_1)^3 + (1-p_1)e_2^3.
\]
Figure~\ref{fig:illustHOC} illustrates the above consensus checking process. We leave more details and full derivations to Appendix~\ref{supp:derivation}. The left-hand side of each above equation is the probability of a particular first-, second-, or third-order consensus pattern of $\widetilde Y$, which could be estimated given the noisy dataset $\widetilde D$.
These consensus patterns encode the high-order information of $\mT$.
Later in Section~\ref{sec:unique_T}, we will prove that given the consensus probability (LHS), the first three order consensus equations we presented above are sufficient to jointly identify a unique solution to $\mT$, which indeed corresponds to the true $\mT$.

\subsection{Estimating \texorpdfstring{$\mT$}{}: The General Form}\label{sec:global}
\vspace{-5pt}

We generalize this idea to classifications with multiple classes.
For a $K$-class classification problem, define
$
\bm p:= [\PP(Y=i), i\in[K]]^\top
$
and 
\vspace{-2pt}
\begin{equation}\label{eq:TS}
    \bm T_r := \mT \cdot \mS_r,~ \forall r \in [K],
\vspace{-2pt}
\end{equation}
where $\mS_r:=[ \ve_{r+1}, \ve_{r+2}, \cdots, \ve_{K}, \ve_{1}, \ve_{2}, \cdots \ve_{r}]$ is a cyclic permutation matrix, and $\ve_{r}$ is the $K\times 1$ column vector of which the $r$-th element is $1$ and $0$ otherwise.
The matrix $\mS_r$ cyclically shifts each column of $\mT$ to its left side by $r$ units.
Similar to the previous binary example, the LHS of the equation is the probability of different distributions of $\widetilde Y$s among each instance and its $2$-NN.
Let $(i+r)_K := [(i+r-1) \mod K] + 1$. For the first-, second-, and third-order consensuses, we can respectively denote them in vector forms as follows ($\forall r \in [K], s \in [K]$).

{\vspace{-15pt}\small\begin{align*}
    \vc^{[1]} &= [\PP(\widetilde Y_1=i),i\in[K]]^\top, \\
    \vc^{[2]}_{r} &= [\PP(\widetilde Y_1=i,\widetilde Y_2=(i+r)_K), i \in [K]]^\top, \\
    \vc^{[3]}_{r, s} \hspace{-2pt} &= \hspace{-2pt}[\PP(\widetilde Y_1=i,\widetilde Y_2=(i+r)_K,\widetilde Y_3=(i+s)_K),  i \in [K]]^\top.
\end{align*}}
Denote by $\circ$ the Hadamard product of two matrices. We now present the system of consensus equations for estimating $\mT$ and $\vp$ in the general form:%
\begin{tcolorbox}[colback=grey!5!white,colframe=grey!5!white]
\vspace{-5pt}
\begin{center}
    \textbf{Consensus Equations}
\end{center}
\vspace{-9pt}
\begin{itemize}\setlength\itemsep{-0.8em}\setlength{\itemindent}{-2em}
    \item First-order ($K$ equations):
    \vspace{-5pt}
    \begin{align}\label{eq:o1}
        \vc^{[1]} := \mT^\top  \vp,
    \end{align}
    \item Second-order ($K^2$ equations): 
    \vspace{-5pt}
    \begin{align}\label{eq:o2}
        \vc^{[2]}_{r} := (\mT\circ \mT_r)^\top \vp, ~r \in [K],
    \end{align}
    \item Third-order ($K^3$ equations):
    \vspace{-5pt}
    \begin{align}\label{eq:o3}
        \vc^{[3]}_{r,s} := (\mT\circ \mT_r \circ \mT_s)^\top  \vp, ~r,s \in [K].
    \end{align}
\end{itemize}
\vspace{-15pt}
\end{tcolorbox}
\vspace{-7pt}

While we leave the full details of derivation to Appendix~\ref{supp:derivation}, we show one second-order consensus below for an example:

{\vspace{-17pt}
\small
\begin{align*}
     & \ve_j^\top \vc^{[2]}_{r} = \PP(\widetilde Y_1 = j,\widetilde Y_2 = (j+r)_K) \\
    \overset{(a)}{=}&\sum_{i\in[K]}\PP(\widetilde Y_1 = j| Y_1 = i)\PP(\widetilde Y_2 = (j+r)_K| Y_2 = i) \PP(Y_1=i) \\
    {=}&\sum_{i\in[K]} T_{i,j} \cdot T_{i,(j+r)_K} \cdot p_i   \overset{(b)}{=} \ve_j^\top (\mT \circ \mT_r)^\top \vp,
\end{align*}
\vspace{-17pt}
}

where equality $(a)$ holds again due to the $2$-NN label clusterability and the conditional independency (similar to binary cases), and equality $(b)$ holds due to $\mT_r[i,j] = T_{i,(j+r)_K}$. 

We note that although there are higher-order consensuses according to this rule, we only consider up to third-order consensuses of $\widetilde Y$ as shown in Eqns.~(\ref{eq:o1})--(\ref{eq:o3}).
For ease of notation, we define two stacked vector-forms for $\vc^{[2]}_{r,s}, \vc^{[3]}_{r,s} $ :
\vspace{-3pt}
\begin{align}\label{eq:long2}
    \vc^{[2]}:&=[(\vc^{[2]}_{r})^\top,\forall r\in[K]]^\top,\\
\label{eq:long3}
    \vc^{[3]}:&=[(\vc^{[3]}_{r,s})^\top,\forall r,s \in [K]]^\top.
\end{align}

\subsection{The \ours{} Estimator}\label{sec:HOCestimator}

Solving the consensus equations requires estimating the consensus probabilities $\vc^{[1]}$, $\vc^{[2]}$, and $\vc^{[3]}$. In this subsection, we will first show the procedures for estimating these probabilities and then formulate an efficient optimization problem for $\mT$ and $\vp$.
To summarize, there are three steps:
\squishlist
\item \textbf{Step 1:} 
Find $2$-NN for each $\bar x_n$ from the noisy dataset $\widetilde D$.
\item \textbf{Step 2:} Compute each $\hat\vc^{[\nu]}$ using $\bar x_n$ and their 2-NN. %
\item \textbf{Step 3:} Formulate the optimization problem in (\ref{eq:problem2}).
\squishend
Denote by $E\subseteq[N]$.
We elaborate on each step as follows.

\vspace{-3pt}
\textbf{Step 1: Find $2$-NN~} 
Given the noisy dataset $\{(x_n,\tilde y_n), n \in E\}$, for each representation $\bar x_n=\vf_{\sf conv}(x_n)$, we can find its $2$-NN $\bar x_{n_1}, \bar x_{n_2}$ as:

{\vspace{-17pt}\small
\[
n_1 = \hspace{-5pt}\argmin_{n'\in E,n'\ne n}~\hspace{-5pt}\textsf{Dist}(\bar x_n, \bar x_{n'}), ~
n_2 = \hspace{-5pt}\argmin_{n'\in E,n'\ne n\ne n_1}\hspace{-5pt}\textsf{Dist}(\bar x_n, \bar x_{n'}),  
\]\vspace{-17pt}}

and the corresponding noisy labels $\tilde y_{n_1}, \tilde y_{n_2}$. $\textsf{Dist}(A,B)$ measures the distance between $A$ and $B$ - we will use $\textsf{Dist}$ as the negative cosine similarity in our experiment.  %

\textbf{Step 2: Empirical mean~} 
Denote by $\BR\{\cdot\}$ the indicator function taking value $1$ when the specified condition is met and $0$ otherwise. Let $E$ be a set of indices and $|E|$ be the number of them.
The probability of each high-order consensus could be estimated by the empirical mean using a particular set of sampled examples in $E$: $\{(\tilde y_n, \tilde y_{n_1}, \tilde y_{n_2}), n\in E\}$ as follows ($\forall i$).

{\vspace{-17pt}
\small
\begin{align}
    {\hat{\vc}}^{[1]}[i] &= \frac{1}{|E|} \sum_{n\in E}\BR{\{\tilde y_n =i\}},\nonumber \\
    {\hat{\vc}}^{[2]}_{r}[i] &= \frac{1}{|E|} \sum_{n\in E}\BR{\{\tilde y_n =i,\tilde y_{n_1}=(i+r)_K\}}, \label{eq:EM_HOC} \\
    {\hat{\vc}}^{[3]}_{r,s}[i] &= \frac{1}{|E|} \sum_{n\in E}\BR{\{\tilde y_n =i,\tilde y_{n_1}=(i+r)_K,\tilde y_{n_2}=(i+s)_K\}}.\nonumber
\end{align}
\vspace{-17pt}}

The motivation of identifying a subset $E$ for the estimators is due to the desired provable convergence to the expectation. Each $3$-tuple in the sample should be independent and identically distributed (i.i.d.) so that each ${\hat{\vc}}^{[\nu]}$ is consistent. However, the existence of nearest neighbors, e.g., when both $n$ and $n_1$ belong to $E$ and $n$ is a $2$-NN of $n_1$, may violate the i.i.d. property of these $3$-tuples.
Denote by
\[
E_3^* = \argmax_{E\subseteq [N]} |E|,~~ \textsf{s.t.}~~ |\{ n, n_1, n_2, \forall n \in E \}| = 3|E|.
\]
Then any subset $E \subseteq E_3^*$ guarantees the i.i.d. property.
Note it is generally time-consuming to find the best $E$.
For an efficient solution (with empirical approximation), we randomly sample $|E|$ center indices from $[N]$ and repeat Step 1 and Step 2 multiple times with different $E$ (as Line~3~--~Line~8 in Algorithm~\ref{algorithm1}).
We will further discuss the magnitude of $|E|$ in Section~\ref{sec:sample_cplxy} and Appendix~\ref{supp:proof_c}.

\textbf{Step 3: Optimization~}
With ${\hat{\vc}}^{[1]}$, ${\hat{\vc}}^{[2]}$, and ${\hat{\vc}}^{[3]}$, 
we formulate the optimization problem in (\ref{eq:problem1}) to jointly solve for $\mT,\vp$.
\vspace{-12pt}
\begin{subequations}\label{eq:problem1}
\begin{align}
\mathop{\sf minimize}\limits_{\mT,\vp} \qquad  &\sum_{\nu=1}^3\|{\hat{\vc}}^{[\nu]} - {{\vc}}^{[\nu]}\|_2 \label{eq:p1_obj} \\
{\sf subject~to} \qquad
& \text{Eqns.~(\ref{eq:TS})~--~(\ref{eq:long3})} \label{eq:p1_var} \\
& p_{i} \ge 0, T_{ij} \ge 0,  i,j\in[K] \label{eq:p1_c1}\\
& \sum_{i\in[K]} p_i = 1, \sum_{j\in[K]} T_{ij} = 1,  i\in[K] \label{eq:p1_c2}.
\end{align}
\end{subequations}
The crucial components in (\ref{eq:problem1}) are:
\squishlist
\item Objective (\ref{eq:p1_obj}): the sum of errors from each order of consensus, where the error is defined in $\ell_2$-norm.
\item Variable definitions (\ref{eq:p1_var}): the closed-form relationship between intermediate variables (such as $\vc^{[\nu]}$ and $\mT_r$) and the optimized variables ($\mT$ and $\vp$). 
\item Constraints (\ref{eq:p1_c1}) and (\ref{eq:p1_c2}): feasibility of a solution.
\squishend

\textbf{Challenges for solving the constrained optimization problem}~ The problem in (\ref{eq:problem1}) is a constrained  
optimization problem with $K(K+1)$ variables, $K(K+1)$ inequality constraints, and $(K+1)$ equality constraints, and it is generally hard to guarantee its convexity. Directly solving this problem using the Lagrangian-dual method may take a long time to converge \cite{boyd2004convex}. %

\textbf{Unconstrained soft approximation}~ Notice that both $\vp$ and each row of $\mT$ are probability measures.
Instead of directly solving for $\mT$ and $\vp$, we seek to relax the constraints by introducing auxiliary and unconstrained variables to represent $\mT$ and $\vp$. %
Particularly, we turn to optimizing variables $\bar\mT \in \mathbb R^{K\times K}$ and $\bar\vp \in \mathbb R^{K}$ that are associated with $\mT$ and $\vp$ by $\mT := \sigma_T(\bar\mT),~ \vp := \sigma_p (\bar \vp)$, where $\sigma_T(\cdot)$ and $\sigma_p(\cdot)$ are softmax functions such that
\begin{equation}\label{eq:softmax}
    T_{ij} := \frac{\exp(\bar T_{ij})}{\sum_{k\in[K]}\exp(\bar T_{ik})}, ~ p_i := \frac{\exp(\bar p_{i})}{\sum_{k\in[K]}\exp(\bar p_{k})}.
\end{equation}
Therefore, we can drop all the constraints in (\ref{eq:problem1}) and focus on solving the unconstrained optimization problem with $K(K+1)$ variables.
Our new optimization problem is given as follows:
\vspace{-10pt}
\begin{subequations}\label{eq:problem2}
\begin{align}
\mathop{\sf minimize}\limits_{\bar\mT, \bar\vp} \qquad  &\sum_{\nu=1}^3\|{\hat{\vc}}^{[\nu]} - {{\vc}}^{[\nu]}\|_2 \label{eq:p2_obj} \\
{\sf subject~to} \qquad
& \text{Eqns.~(\ref{eq:TS})~--~(\ref{eq:long3}), Eqn.~(\ref{eq:softmax})}.  \label{eq:p2_var}
\end{align}
\end{subequations}
Equations in (\ref{eq:p2_var}) are presented only for a clear objective function.
Given the solution of problem (\ref{eq:problem2}), we can calculate $\mT$ and $\vp$ according to Eqn.~(\ref{eq:softmax}).
\rev{Note the search space of $\mT$ before and after soft approximation differs only in corner cases (before: $T_{ij}\ge 0$, after: $T_{ij}>0$).  
For each original and non-corner $\mT$, there exists a soft approximated $\mT$ that leads to the same transition probabilities. Thus the soft approximation preserves the property of $\mT$, e.g. the uniqueness in Theorem~\ref{thm:unique}.}
Algorithm~\ref{alg:Test} summarizes our High-Order-Consensus (\ours{}) estimator.
\begin{algorithm}[tb]
   \caption{The \ours{} Estimator}
   \label{alg:Test}
\begin{algorithmic}[1]
   \STATE {\bfseries Input:} Rounds: $G$. Sample size: $|E|$. Noisy dataset: $\widetilde{D}=\{( x_n, \tilde y_n)\}_{n\in  [N]}$. Representation extractor: $\vf_{\sf conv}$.
  \STATE {\bfseries Initialization:} Set ${\hat{\vc}}^{[1]}$, ${\hat{\vc}}^{[2]}$,  ${\hat{\vc}}^{[3]}$ to $0$. 
  Extract representations $x_n\leftarrow \vf_{\sf conv}(x_n), \forall n\in[N]$.
  $\bar\mT = K\mI - \1 \1^\top$. $\bar\vp = \1/K$.
  \algcom{// $\mI$: identity matrix, $\1$: all-ones column vector.}
    \REPEAT \label{line:rndsmp_0}
    \STATE $E\leftarrow$ \texttt{RndSmp}$([N],|E|)$;
    \algcom{// sample $|E|$ center indices}
    \\
    \STATE $\{(\tilde y_n,\tilde y_{n_1},\tilde y_{n_2}), n\in[E]\}\leftarrow$ \texttt{Get2NN}$(\widetilde{D},E)$; \\
    \hfill \algcom{// find the noisy labels of the $2$-NN of $x_n,n\in[E]$}\\
    \STATE $({\hat{\vc}}^{[1]}_{\sf{tmp}}$, ${\hat{\vc}}^{[2]}_{\sf{tmp}}$,  ${\hat{\vc}}^{[3]}_{\sf{tmp}})\leftarrow$  \texttt{CountFreq}($E$) \algcom{// as Eqn.~(\ref{eq:EM_HOC})}
    \STATE ${\hat{\vc}}^{[\nu]} \leftarrow {\hat{\vc}}^{[\nu]} + {\hat{\vc}}^{[\nu]}_{\sf{tmp}}, \nu \in \{1,2,3\}$;
    \UNTIL{$G$ times} \label{line:rndsmp_1}
    \STATE ${\hat{\vc}}^{[\nu]} \leftarrow {\hat{\vc}}^{[\nu]} / G, \nu \in \{1,2,3\}$; \hfill \algcom{// estimate $\vc^{[\nu]}$ $G$ times}
    \STATE Solve the unconstrained problem %
    in (\ref{eq:problem2}) with ${(\hat{\vc}}^{[1]},{\hat{\vc}}^{[2]},{\hat{\vc}}^{[3]})$ by gradient decent, get $\bar\mT$ and $\bar\vp$%
    \STATE {\bfseries Output:} Estimates $\hat\mT \leftarrow \sigma_T(\bar\mT),~ \hat\vp \leftarrow \sigma_p (\bar \vp)$.
\end{algorithmic}
\end{algorithm}
\vspace{-10pt}

\subsection{Flexible Extensions to Instance-Dependent Noise}\label{sec:IDN}

Algorithm~\ref{alg:Test} provides a generically applicable and light tool for fast estimation of $\mT$.
The flexibility makes it possible to be applied to more sophisticated instance-dependent label noise.
We briefly discuss possible applications to estimating the local noise transition matrix $\mT(X)$.

\textbf{Locally homogeneous label noise}~
Intuitively, by considering a local dataset in which every representation shares the same $\mT(X)$, the method in Section~\ref{sec:global} can then be applied locally to estimate the local $\mT(X)$. %
Specially, %
using a ``waypoint" $\bar x_n$, we build a local dataset $\widetilde{{D}}_n$ that includes the $M$-NN of $\bar x_n$, i.e., $\widetilde D_n = \{(x_n,\tilde{y}_n)\} \cup \{(x_{n_i},\tilde y_{n_i}), \forall i\in[M]\}$, where $\{n_i, i\in[M]\}$ are the indices of the $M$-NN of $\bar x_n$. We introduce the following definitions:
\begin{defn}[$M$-NN noise clusterability]\label{ap:sameT}
We call $\widetilde D_n$ satisfies $M$-NN noise clusterability if the $M$-NN of $\bar x_n$ have the same noise transition matrix as $x_n$, i.e., $\mT(x_n) = \mT(x_{n_i}), \forall i \in [M]$. %
\end{defn}
\begin{defn}[$(H,M)$-coverage]
We call $\widetilde{D}$ satisfies $(H,M)$-coverage if there exist $H$ instances $\bar x_{h(n)}, n\in[H]$ such that $\widetilde{D} = \cup_{n=1}^H \widetilde{D}_{h(n)}$, where each $\widetilde{D}_{h(n)}$ satisfies $M$-NN noise clusterability. %
\end{defn}

Note Dentition~\ref{ap:sameT} focuses on the clusterability of noise transition matrices, which is different from the clusterability of the true classes of labels.
When $M$-NN noise clusterability holds for $\bar x_n$, the label noise in local dataset $\widetilde{D}_n$ is effectively homogeneous.
If $\widetilde D$ further satisfies $(H,M)$-coverage, we can divide the training data $\widetilde D$ to $H$ local sub-datasets $\widetilde{D}_{h(n)}, n\in[H]$ and separately apply Algorithm~\ref{alg:Test} on each of them.
The local estimates allow us to apply loss correction separately using different $\mT(X)$ at different parts of the training data. 
\rev{Besides, when there is no $M$-NN noise clusterability, we may require knowing properly constructed sub-spaces to separate the data, with each part of them sharing similar noise rates \cite{xia2020extended,xia2020parts}.
We leave more detailed discussions in Appendix~\ref{supp:localT}.}

%% file: src/guarantee.tex
\section{Theoretical Guarantees}
We will prove that our consensus equations are sufficient for estimating a unique $\mT$, and show the advantage of our approach in terms of a better sample complexity than the anchor point approach. 

\subsection{Uniqueness of Solution}\label{sec:unique_T}
Before formally presenting the uniqueness guarantee, we introduce two assumptions as we will need.

\begin{ap}[Nonsingular $\mT$]\label{ap:invT}
The noise transition matrix is non-singular, i.e., ${\sf Rank}(\mT) = K$.
\end{ap}

\begin{ap}[Informative $\mT$]\label{ap:informative}
The diagonal elements of $\mT$ are dominant, i.e., $T_{ii} > T_{ij}, \forall i \in [K], j\in[K], j\ne i$.
\end{ap}

Assumption~\ref{ap:invT} is commonly made in the literature and ensures the effect of label noise is invertible \cite{van2017theory}. Assumption~\ref{ap:informative} characterizes a particular permutation of row vectors in $\mT$ \cite{liu2020surrogate}. See more discussions on their feasibility in Appendix~\ref{supp:feasibilityAP}.
The uniqueness is formally stated in Theorem~\ref{thm:unique}. \rev{The proof is sketched at the end of main paper and detailed in Appendix~\ref{supp:proof_unique}.}
\begin{thm}\label{thm:unique}
When $\widetilde D$ satisfies the $2$-NN label clusterability and $\mT$ is nonsingular and informative, with a perfect knowledge of $\vc^{[\nu]}, \nu=1,2,3$, the solution of consensus equations (\ref{eq:o1})~--~(\ref{eq:o3}) returns the true $\mT$ \underline{uniquely}.
\end{thm}

\textbf{Challenges}~
Proving Theorem~\ref{thm:unique} is challenging due to:
1) The coupling effect between $\mT$ and $\vp$ makes the structure of solution $\mT$ unclear;
2) Naively replacing $\vp$, e.g., using $\vp = (\mT^\top)^{-1}\vc^{[1]}$, will introduce matrix inverse, which cannot be canceled with the Hadamard product; 3) A system of third-order equations with $K^2$ variables will have up to $3^{K^2}$ solutions and the closed-form is not explicit.

\textbf{Local estimates}~
Our next corollary~\ref{cor:IDN_unique} extends Theorem~\ref{thm:unique} to local datasets, when $\mT$ can be heterogeneous. 
\begin{cor}\label{cor:IDN_unique}
When $\widetilde{D}$ satisfies $(H,M)$-coverage, each $ \widetilde{D}_{h(n)}$ satisfies 2-NN label clusterability, and
$\mT(x_{h(n)})$ is nonsingular and informative,
with a perfect knowledge of the local $\vc^{[\nu]}, \nu=1,2,3$,
the solution of consensus equations~(\ref{eq:o1})~--~(\ref{eq:o3}) is unique and recovers $\mT(x_{h(n)})$.
\end{cor}

\subsection{Sample Complexity}\label{sec:sample_cplxy}

We next show that with the estimates  $\hat\vc^{[1]}$, $\hat\vc^{[2]}$, and $\hat\vc^{[3]}$, \ours{} returns a reasonably well solution. 

Recall that, in Section~\ref{sec:HOCestimator}, Step 2 requires a particular $E\subseteq E^*_3$ to guarantee the i.i.d. property of the sample $\{(\tilde y_n, \tilde y_{n_1}, \tilde y_{n_2}), n\in E\}$.
For a tractable sample complexity, we focus on a particular dataset $\widetilde D$ and feature extractor $\vf_{\sf conv}$ such that 1) $|E_3^*|=\Theta(N)$  %
and 2) $T_{ij} = \frac{1-T_{ii}}{K-1}, \forall j\ne i, i\in[N],j\in[N]$.
\rev{Supposing each tuple is drawn from non-overlapping balls, condition 1) is satisfied when the number of these non-overlapping balls covering the representation space is $\Theta(N)$.  See Appendix~\ref{supp:feasibleE} for a detailed example when the representations are uniformly distributed.}
Lemma~\ref{lem:sample_c} shows the error upper bound of our estimates $\hat \vc^{[\nu]}, \nu = 1,2,3.$
See Appendix~\ref{supp:proof_c} for the proof. %

\begin{lem}\label{lem:sample_c}
With probability $1\hspace{-1pt}-\hspace{-1pt}\delta$, $\forall \nu,l,$ the estimation error $|\hat \vc^{[\nu]}[l] \hspace{-1pt}-\hspace{-1pt} \vc^{[\nu]}[l]|$ is bounded at the order of $O(\sqrt{\ln(1/\delta)/N})$.
\end{lem}

Lemma~\ref{lem:sample_c} is effectively the sample complexity of estimating $|E_3^*|$ i.i.d. random variables by the sample mean. %
Due to assuming a uniform diagonal $\mT$, we only need to consider the estimation error of $\hat T_{ii}$.
For each $i\in[K]$, see the result in Theorem~\ref{thm:sample} and the proof in Appendix~\ref{supp:proof_sample}.

\begin{thm}\label{thm:sample}
When $ T_{ii} > \frac{1-\PP(Y=i)+(K-1) \PP(\widetilde Y=i)}{K(K-1)\PP(Y=i)}$, w.p. $1-2\delta$, $|\hat T_{ii}-T_{ii}|$ is bounded at the order of $O(\sqrt{{\ln(1/\delta)}/{N}})$.
\end{thm}

Theorem~\ref{thm:sample} indicates the sample complexity of our solution has the same order in terms of $N$ compared to a standard empirical mean estimation in Lemma~\ref{lem:sample_c}. Remark~\ref{rmk:compare} shows our approach is better than using a set of anchor points in the sample complexity.

\begin{rmk}[Comparison]\label{rmk:compare}
The methods based on anchor points estimate $\mT$ with $N_{\text{AC}}<N$ ($N_{\text{AC}} \ll N$ in many cases) anchor points. Thus w.p. $1-\delta$, the estimation error is at the order of $O(\sqrt{{\ln(1/\delta)}/{N_{\text{AC}}}})$.
\end{rmk}

%% file: src/exp.tex
\begin{figure*}
    \centering
    \includegraphics[width=0.81\textwidth]{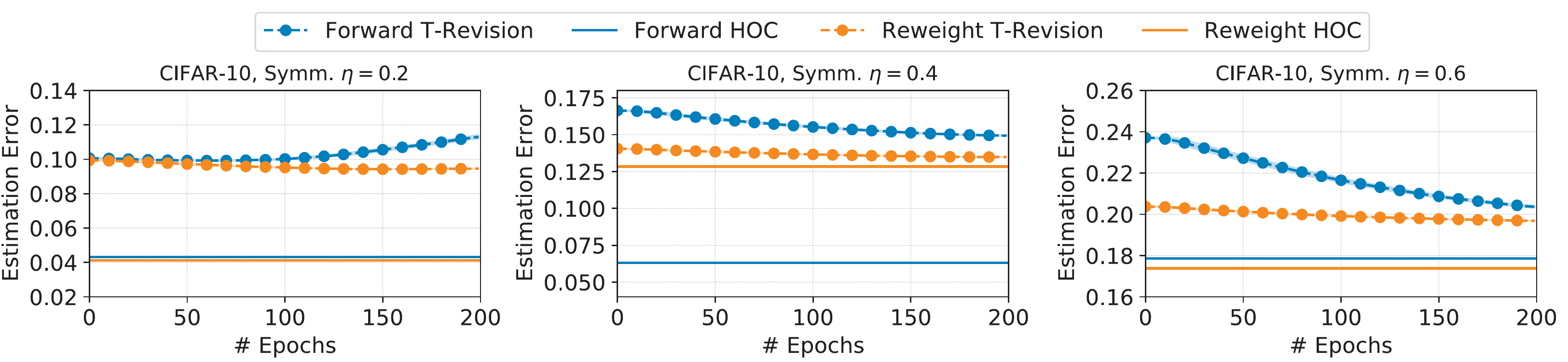}
    \includegraphics[width=0.81\textwidth]{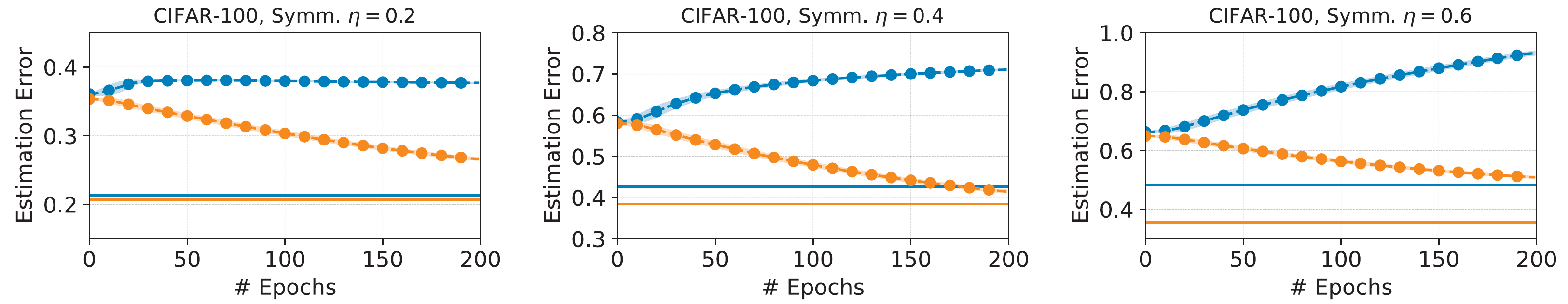}
    \vspace{-5pt}
    \caption{Comparison of estimation errors of $\mT$ given by T-Revision \texorpdfstring{\cite{xia2019anchor}}{} and our \ours{} estimator. The error is measured by the matrix $L_{1,1}$-norm with a normalization factor $K$, i.e. $\|\hat \mT - \mT\|_{1,1}/K$. Forward: Using the forward corrected loss \texorpdfstring{\cite{patrini2017making}}. Reweight: Using the reweighted loss \cite{liu2015classification}. Symmetric noise is applied.}
    \label{fig:est_err}
\end{figure*}
\section{Experiments}

\begin{table*}[!t]
		\caption{The best epoch (clean) test accuracy (\%) with synthetic label noise.}
		\begin{center}
		\scalebox{.8}{{\begin{tabular}{c|cccccc} 
				\hline 
				 \multirow{2}{*}{Method}  & \multicolumn{3}{c}{\emph{Inst. CIFAR-10} } & \multicolumn{3}{c}{\emph{Inst. CIFAR-100} } \\ 
				 & $\eta = 0.2$&$\eta = 0.4$&$\eta = 0.6$ & $\eta = 0.2$&$\eta = 0.4$&$\eta = 0.6$\\
				\hline\hline
			     CE (Standard)  &85.66$\pm$0.62 & 76.89$\pm$0.93 &  60.29$\pm$1.17 &  57.26$\pm$1.33 & 41.33$\pm$0.89 & 25.08$\pm$1.85 \\
 				 Peer Loss \cite{liu2019peer} &89.52$\pm$0.22 & 83.44$\pm$0.30 & 75.15$\pm$0.82  & 61.13$\pm$0.48 & 48.01$\pm$0.12 & 33.00$\pm$1.47\\
				 $L_{\sf DMI}$ \cite{xu2019l_dmi}  &88.67$\pm$0.70 & 83.65$\pm$1.13 & 69.82$\pm$1.72 & 57.36$\pm$1.18 & 43.06$\pm$0.97 & 26.13$\pm$2.39\\
				 $L_{q}$ \cite{zhang2018generalized}   & 85.66$\pm$1.09 & 75.24$\pm$1.07 & 61.30$\pm$3.35  & 56.92$\pm$0.24 & 40.17$\pm$1.52 & 25.58$\pm$3.12\\
				 Co-teaching \cite{han2018co} & 88.84$\pm$0.20 & 72.61$\pm$1.35 & 63.76$\pm$1.11 & 43.37$\pm$0.47 & 23.20$\pm$0.44 & 12.43$\pm$0.50\\
				 Co-teaching+ \cite{yu2019does}  & 89.82$\pm$0.39 & 73.44$\pm$0.38 & 63.61$\pm$1.78 & 41.62$\pm$1.05 & 24.73$\pm$0.85 & 12.25$\pm$0.35\\
				JoCoR \cite{wei2020combating} & 88.82$\pm$0.20 & 71.13$\pm$1.94 & 63.88$\pm$2.05 & 44.55$\pm$0.62 & 23.92$\pm$0.32 & 13.05$\pm$1.10\\
				Forward \cite{patrini2017making} & 87.87$\pm$0.96 & 79.81$\pm$2.58 & 68.32$\pm$1.68  & 57.69$\pm$1.55 & 42.62$\pm$0.92 & 27.35$\pm$3.42\\
				T-Revision \cite{xia2019anchor} & \textbf{90.31$\pm$0.37} & 84.99$\pm$0.81 & 72.06$\pm$3.40 & 58.00$\pm$0.20 & 40.01$\pm$0.32 & 40.88$\pm$7.57\\
				\ours{} Global & {89.71$\pm$0.51} & {84.62$\pm$1.02} &  {70.67$\pm$3.38} & \textbf{68.82$\pm$0.26} & \textbf{62.29$\pm$1.11} & \textbf{52.96$\pm$1.85}\\
				\ours{} Local & \textbf{90.03$\pm$0.15} & \textbf{85.49$\pm$0.80} &  \textbf{77.40$\pm$0.47} & 67.47$\pm$0.85 & 61.20$\pm$1.04 & 49.84$\pm$1.81\\
			   \hline
			\end{tabular}}}
		\end{center}
		\vspace{-7pt}
		\label{table:cifar-inst}
\end{table*}

We present experiment settings as follows.

\textbf{Datasets and models~}
\ours{} is evaluated on three benchmark datasets: CIFAR-10, CIFAR-100 \cite{krizhevsky2009learning} and Clothing1M \cite{xiao2015learning}.
For the standard training step, we use ResNet34 for CIFAR-10 and CIFAR-100, and ResNet50 for Clothing1M.
The representations come from the outputs before the final fully-connected layer of ResNet34/50.
The distance between different representations is measured by the negative cosine similarity.

\textbf{Noise type~}
\ours{} is tested on both synthetic label noise and real-world human label noise.
The synthetic label noise includes two regimes: \emph{symmetric} noise and \emph{instance-dependent} noise.
For both regimes, the noise rate $\eta$ is the overall ratio of instances with a corrupted label in the whole dataset.
The symmetric noise is generated by randomly flipping a clean label to the other possible classes w.p. $\eta$ \citep{xia2019anchor}.
The basic idea of generating instance-dependent noise is to randomly generate one vector for each class ($K$ vectors in total) and project each incoming  feature onto these $K$ vectors \cite{xia2020parts}. The label noise is added by jointly considering the clean label and the projection results. See Appendix~\ref{sec:instance_noise_gen} for more details.
The \emph{real-world human noise} comes from human annotations.
Particularly, for the $50,000$ training images in CIFAR-10, we \emph{re-collect} human annotations\footnote{We only collect one annotation for each image with a cost of \textcent $10$ per image.} from Amazon Mechanical Turk (MTurk) in February 2020.
For the Clothing1M dataset, we train on 1 million noisy training instances reflecting the real-world human noise.

\subsection{Performance of Estimating \texorpdfstring{$\mT$}{}}

We compare \ours{} with T-revision \cite{xia2019anchor} following the flow: 1) Estimation $\rightarrow$ 2) Training $\rightarrow$ 3) Revision.
For a fair comparison, we follow their training framework and parameter settings to get representations. Particularly, we obtain the same model as the one that T-revision adopts before revision.
As illustrated in Figure~\ref{fig:est_err}, compared with the dynamical revision adopted in T-revision, \ours{} does not need to change or adapt in different epochs and still achieves lower estimation errors no matter the model is trained with forward corrected loss or reweighted loss.

\subsection{Performance of Classification Accuracy}

\begin{table}[tb]
	\caption{The best epoch test accuracy (\%) with human noise.
	}
	\begin{center}
	\scalebox{.75}{{
		\begin{tabular}{c|cc} 
			\hline 
			Method & Clothing1M & Human CIFAR-10 \\
			\hline \hline 
			CE (standard) &  68.94 & 83.50\\
			\SPL{} \cite{sieve2020} & 73.24 & 89.98\\
			$L_{\sf DMI}$ \cite{xu2019l_dmi} & 72.46 & 86.33\\
			Co-teaching \cite{han2018co}  &69.21 & 90.39 \\
			JoCoR \cite{wei2020combating} &70.30 & 90.10 \\
			Forward  \cite{patrini2017making} & 70.83 & 86.82\\
			PTD-R-V\cite{xia2020parts} & 71.67 & 85.92 \\
			\ours{}  & \textbf{73.39} & \textbf{90.62}\\
			\hline 
		\end{tabular}
		}}
	\end{center}
	\vspace{-15pt}
	\label{table:c1m}
\end{table}

To test the classification performance, we adopt the flow: 1) Pre-training $\rightarrow$ 2) Global Training $\rightarrow$ 3) Local Training.
Our \ours{} estimator is applied once at the beginning of each above step.
In Stage-1, we load the standard ResNet50 model pre-trained on ImageNet to obtain basic representations.
At the beginning of Stage-2 and Stage-3, we use the representations given by the current model.
All experiments are repeated three times.
\emph{\ours{} Global} only employs one global $\mT$ with $G=50$ and $|E|=15k$ as inputs of Algorithm~\ref{algorithm1}. \emph{\ours{} Local} uses $300$ local matrices %
($250$-NN noise clusterability, $G=30$, $|E|=100$) for CIFAR-10 and $5$ local matrices ($10k$-NN noise clusterability, $G=30$, $|E|=5k$) for CIFAR-100.\footnote{Our unconstrained transformation provides much better convergence such that running \ours{} Local on CIFAR will at most double the running time of a standard training with CE.}
See more details in Appendix~\ref{supp:exp_set}.
Without sophisticated learning techniques, we simply feed the estimated transition matrices given by \ours{} into \emph{forward loss correction} \cite{patrini2017making}.
We report the performance on synthetic instance-dependent label noise in Table~\ref{table:cifar-inst} and real-world human-level label noise in Table~\ref{table:c1m}.
Comparing with these baselines (with similar data augmentations), both global estimates and local estimates given by \ours{} achieve satisfying performance, and the local estimates indeed provide sufficient performance improvement on CIFAR-10.
When there are $100$ classes, $\mT$ contains $10k$ variables thus local estimates with only $10k$ instances may not be accurate, which leads to a slight performance drop in \ours{} Local on CIFAR-100 (but it still outperforms other methods).

\textbf{Real human-level noise}~ On CIFAR-10 with our self-collected human-level noisy labels, \ours{} achieves a $0.097$ estimation error in the global $\mT$ and a $0.110\pm 0.027$ error in estimating $300$ local transition matrices. See more details in Appendix~\ref{supp:est_error_local}.

\rev{
\subsection{Feasibility of $2$-NN label clusterability}\label{sec:feasibility}

We show the ratio of feasible $2$-NN tuples in Table~\ref{table:2nn}.
One $2$-NN tuple is called feasible if $\bar x_n$ and its $2$-NN belong to the same true class. The feature extractors are obtained from overfitting CIFAR-10/100 with different noise levels. For example, \emph{CIFAR-10 Inst.} $\eta=0.2$ indicates that we use the standard CE loss to train ResNet34 on CIFAR-10 with $20\%$ instance-dependent label noise. The convolution layers when the model approaches nearly $100\%$ training accuracy are selected as the feature extractor $\vf_{\sf conv}(X)$. Table~\ref{table:2nn} shows, with a standard feature extractor, there are more than $2/3$ of the feasible $2$-NN tuples in most cases. Besides, reducing the sample size from $50k$ to $5k$ will not substantially reduce the ratio of feasible $2$-NN tuples.

\begin{table}[!t]
		\caption{The ratio of feasible $2$-NN tuples with different feature extractors. $|E|=5k$: Sample $5k$ examples from the whole dataset in each round, and average over $10$ rounds. $|E|=50k$: Check the feasibility of all $2$-NN tuples.}
		\begin{center}
		\scalebox{.75}{{\begin{tabular}{c|cccccc} 
				\hline 
				 \multirow{2}{*}{Feature Extractor}  & \multicolumn{3}{c}{\emph{CIFAR-10} } & \multicolumn{3}{c}{\emph{CIFAR-100} } \\ 
				 & $|E|=5k$&$|E|=50k$ &  $|E|=5k$&$|E|=50k$\\
				\hline\hline
				\emph{Clean} & 99.99 & 99.99 & 99.88 & 99.90 \\
			     \emph{Inst.} $\eta=0.2$  & 87.88 &  89.06  &  82.82 &  84.33 \\
 				 \emph{Inst.} $\eta=0.4$ & 78.15   & 79.85 & 64.88 & 68.31\\
			   \hline
			\end{tabular}}}
		\end{center}
		\vspace{-7pt}
		\label{table:2nn}
\end{table}
}

%% file: src/conclusion.tex
\section{Conclusions}
This paper has proposed a new and flexible estimator of the noise transition matrix relying on the first-, second-, and third-order consensuses checking among an example and its' $2$-NN's noisy labels.
Future directions of this work include extending our estimator to collaborate with other learning with noisy label techniques. We are also interested in developing algorithms to identify critical masses of instances that share similar noise rates such that our estimator can be applied to local estimation more efficiently. %

\rev{
\section*{Proof Sketch for Theorem~\ref{thm:unique}}

The high-level idea of the proof is to connect the Hadamard products to matrix products, and prove that any linear combination of two or more rows of $\mT$ does not exist in $\mT$.

\textbf{Step I: Transform the second-order equations.}~
By exploiting the relation between Hadamard products and matrix products, the second-order equations can be transformed to ${\mT}^\top \mD_{\vp} {\mT}= \mT_{\dag}$,
where $\mT_\dag$ is fixed given $\vc^{[2]}_r, \forall r\in[K]$, and $\mD_{\vp}$ is a diagonal matrix with $\vp$ as its main diagonals,

\textbf{Step II: Transform the third-order equations.}~
Following the idea in Step I, we can also transform the third-order equations to $(\mT \circ \mT_s)   = \mT\mT_{\dag}^{-1}\mT_{\ddag,s}^\top, \forall s\in[K]$,   where $\mT_{\ddag,s}$ is fixed given $\vc^{[3]}_{r,s}, \forall r,s$.

\textbf{Step III: From matrices to vectors}~
We analyze the rows $\vu^\top$ of $\mT$ and transform the equations in Step II to (e.g. $s=0$) $\mA\vu = \vu \circ  \vu$, where $\mA = \mT_{\ddag}(\mT_{\dag}^{-1})^\top$.
Then we need to find the number of feasible vectors $\vu$.

\textbf{Step IV: Construct the $(K+1)$-th vector}~
When $\mT$ is non-singular, we prove the $(K+1)$-th solution $\vu_{K+1}$ must be identical $\vu_k, k\in[K]$.

\textbf{Wrapping-up: Unique $\mT$}~
Step IV shows $\mT$ only contains $K$ different feasible rows. The informativeness of $\mT$ ensures the unique order of these $K$ rows. Thus $\mT$ is unique.

}

%% file: src/appendix.tex
\appendix
\onecolumn

{\LARGE \bf Appendix}

The Appendix is organized as follows. 
\begin{itemize}
    \item Section~\ref{supp:derivation} presents the detailed examples and derivations of consensus equations.
    \item Section~\ref{supp:theorem} includes proofs and other details about our theoretical results. Particularly,
    \begin{itemize}
        \item Section~\ref{supp:proof_unique} proves the uniqueness of $\bm T$.
        \item Section~\ref{supp:feasibleE} justifies the feasibility of assumption $|E_3^*| = \Theta(N)$
        \item Section~\ref{supp:proof_c} shows the proof for Lemma~\ref{lem:sample_c}
        \item Section~\ref{supp:proof_sample} shows the proof for Theorem~\ref{thm:sample}.
    \end{itemize}
    \item Section~\ref{supp:discuss} presents more discussions, e.g., the soft $2$-NN label clusterability, more details on local $\bm T(X)$, and the feasibility of our Assumption~1~\&~2 to guarantee the uniqueness of $\bm T$.
    \item Section~\ref{supp:exp_set} shows more experimental settings and results.  
\end{itemize}

\section{Derivation of Consensus Equations}\label{supp:derivation}

For the first-order consensuses, we have
\begin{align*}
     &\PP(\widetilde Y_1 = j_1)=\sum_{i\in[K]}\PP(\widetilde Y_1 = j_1| Y_1=i) \PP(Y_1=i).
\end{align*}

For the second-order consensuses, we have
\begin{align*}
     &\PP(\widetilde Y_1 = j_1,\widetilde Y_2 = j_2) \\
     =&\sum_{i\in[K]}\PP(\widetilde Y_1 = j_1,\widetilde Y_2 = j_2| Y_1= i, Y_2 = i) \PP(Y_1=Y_2=i)\\
    \overset{(a)}{=}&\sum_{i\in[K]}\PP(\widetilde Y_1 = j_1,\widetilde Y_2 = j_2| Y_1= i, Y_2 = i)\cdot \PP(Y_1=i) \\
    \overset{(b)}{=}&\sum_{i\in[K]}\PP(\widetilde Y_1 = j_1| Y_1 = i)\cdot\PP(\widetilde Y_2 = j_2| Y_2 = i)\cdot \PP(Y_1=i),
\end{align*}
where equality $(a)$ holds due to the $2$-NN label clusterability, i.e., $Y_1 = Y_2 (=Y_3)$ w.p. 1, and equality $(b)$ holds due to the conditional independency between  $\widetilde Y_1$ and $\widetilde Y_2$ given their clean labels.

For the third-order consensuses, we have
\begin{align*}
     &\PP(\widetilde Y_1 = j_1,\widetilde Y_2 = j_2,\widetilde Y_3 = j_3) \\
    =&\sum_{i\in[K]}\PP(\widetilde Y_1 = j_1,\widetilde Y_2 = j_2,\widetilde Y_3 = j_3| Y_1=i, Y_2=i, Y_3=i) \PP(Y_1=Y_2=Y_3=i) \\
    \overset{(a)}{=}&\sum_{i\in[K]}\PP(\widetilde Y_1 = j_1,\widetilde Y_2 = j_2,\widetilde Y_3 = j_3| Y_1= i, Y_2 = i, Y_3=i) \PP(Y_1=i) \\
    \overset{(b)}{=}&\sum_{i\in[K]}\PP(\widetilde Y_1 = j_1| Y_1 = i)\PP(\widetilde Y_2 = j_2| Y_2 = i)\PP(\widetilde Y_3 = j_3| Y_3 = i) \PP(Y_1=i).
\end{align*}
where equality $(a)$ holds due to the $3$-NN label clusterability, i.e., $Y_1 = Y_2 = Y_3$ w.p. 1, and equality $(b)$ holds due to the conditional independency between $\widetilde Y_1$, $\widetilde Y_2$  and $\widetilde Y_3$ given their clean labels. 

With the above analyses, there are $2$ first-order equations, 
\begin{equation*}
    \begin{split}
        \PP(\widetilde Y_1=1) &= p_1(1-e_1) + (1-p_1)e_2, \\
        \PP(\widetilde Y_1=2) &= p_1 e_1 + (1-p_1)(1-e_2). 
    \end{split}
\end{equation*}
There are $4$ second-order equations for different combinations of $\widetilde{Y}_1,\widetilde{Y}_2$, e.g.,
\begin{equation*}
    \begin{split}
        \PP(\widetilde Y_1=1, \widetilde Y_2=1) &= p_1(1-e_1)^2 + (1-p_1)e_2^2, \\
        \PP(\widetilde Y_1=1, \widetilde Y_2=2) &= p_1(1-e_1)e_1 + (1-p_1)e_2(1-e_2), \\
        \PP(\widetilde Y_1=2, \widetilde Y_2=1) &= p_1(1-e_1)e_1 + (1-p_1)e_2(1-e_2), \\
        \PP(\widetilde Y_1=1, \widetilde Y_2=1) &= p_1 e_1^2 + (1-p_1)(1-e_2)^2.
    \end{split}
\end{equation*}

There are $8$ third-order equations for different combinations of $\widetilde{Y}_1,\widetilde{Y}_2, \widetilde{Y}_3$, e.g.,
\begin{equation*}
    \begin{split}
        \PP(\widetilde Y_1=1, \widetilde Y_2=1, \widetilde Y_3=1) &= p_1(1-e_1)^3 + (1-p_1)e_2^3, \\
        \PP(\widetilde Y_1=1, \widetilde Y_2=1, \widetilde Y_3=2) &= p_1(1-e_1)^2 e_1 + (1-p_1) e_2^2 (1-e_2), \\
        \PP(\widetilde Y_1=1, \widetilde Y_2=2, \widetilde Y_3=1) &= p_1(1-e_1)^2 e_1 + (1-p_1) e_2^2 (1-e_2), \\
        \PP(\widetilde Y_1=1, \widetilde Y_2=2, \widetilde Y_3=2) &= p_1(1-e_1)e_1^2 + (1-p_1) e_2 (1-e_2)^2, \\
        \PP(\widetilde Y_1=2, \widetilde Y_2=1, \widetilde Y_3=1) &= p_1(1-e_1)^2 e_1 + (1-p_1) e_2^2 (1-e_2), \\
        \PP(\widetilde Y_1=2, \widetilde Y_2=1, \widetilde Y_3=2) &= p_1(1-e_1)e_1^2 + (1-p_1) e_2 (1-e_2)^2, \\
        \PP(\widetilde Y_1=2, \widetilde Y_2=2, \widetilde Y_3=1) &= p_1(1-e_1)e_1^2 + (1-p_1) e_2 (1-e_2)^2, \\
        \PP(\widetilde Y_1=2, \widetilde Y_2=2, \widetilde Y_3=2) &= p_1 e_1^3 + (1-p_1)(1-e_2)^3.
    \end{split}
\end{equation*}

For a general $K$-class classification problem, we show one first-order consensus below:
\begin{align*}
     & \ve_j^\top \vc^{[1]} = \PP(\widetilde Y_1 = j) \\
    {=}&\sum_{i\in[K]}\PP(\widetilde Y_1 = j| Y_1 = i) \PP(Y_1=i) \\
    {=}&\sum_{i\in[K]} T_{ij}  \cdot p_i   {=} \ve_j^\top \mT^\top \vp.
\end{align*}
The second-order consensus follows the example below:
\begin{align*}
     & \ve_j^\top \vc^{[2]}_{r} = \PP(\widetilde Y_1 = j,\widetilde Y_2 = (j+r)_K) \\
    \overset{(a)}{=}&\sum_{i\in[K]}\PP(\widetilde Y_1 = j| Y_1 = i)\PP(\widetilde Y_2 = (j+r)_K| Y_2 = i) \PP(Y_1=i) \\
    {=}&\sum_{i\in[K]} T_{i,j} \cdot T_{i,(j+r)_K} \cdot p_i   \overset{(b)}{=} \ve_j^\top (\mT \circ \mT_r)^\top \vp,
\end{align*}
where equality $(a)$ holds again due to the $2$-NN label clusterability the conditional independency (similar to binary cases), and equality $(b)$ holds due to $\mT_r[i,j] = T_{i,(j+r)_K}$. 
We also show one third-order consensus below:
\begin{align*}
     & \ve_j^\top \vc^{[3]}_{r} = \PP(\widetilde Y_1 = j,\widetilde Y_2 = (j+r)_K, \widetilde Y_3 = (j+s)_K) \\
    \overset{(a)}{=}&\sum_{i\in[K]}\PP(\widetilde Y_1 = j| Y_1 = i)\PP(\widetilde Y_2 = (j+r)_K| Y_2 = i)\PP(\widetilde Y_3 = (j+s)_K| Y_3 = i) \PP(Y_1=i) \\
    {=}&\sum_{i\in[K]} T_{i,j} \cdot T_{i,(j+r)_K} \cdot T_{i,(j+s)_K} \cdot p_i   \overset{(b)}{=} \ve_j^\top (\mT \circ \mT_r \circ \mT_s)^\top \vp,
\end{align*}
where equality $(a)$ holds again due to the $3$-NN label clusterability the conditional independency (similar to binary cases), and equality $(b)$ holds due to $\mT_r[i,j] = T_{i,(j+r)_K}$, $\mT_s[i,j] =  T_{i,(j+s)_K}$.

\section{Theoretical Guarantees}\label{supp:theorem}

\subsection{Uniqueness of \texorpdfstring{$\mT$}{}}\label{supp:proof_unique}

We need to prove the following equations have a unique solution when $\mT$ is non-singular and informative.
\begin{tcolorbox}[colback=grey!5!white,colframe=grey!5!white]
\begin{center}
    \textbf{Consensus Equations}
\end{center}
\begin{itemize}\setlength\itemsep{-0.5em}\setlength{\itemindent}{-2em}
    \item First-order ($K$ equations):
    \begin{align*}
        \vc^{[1]} := \mT^\top  \vp,
    \end{align*}
    \item Second-order ($K^2$ equations): 
    \begin{align*}
        \vc^{[2]}_{r} := (\mT\circ \mT_r)^\top \vp, ~r \in [K],
    \end{align*}
    \item Third-order ($K^3$ equations):
    \begin{align*}
        \vc^{[3]}_{r,s} := (\mT\circ \mT_r \circ \mT_s)^\top  \vp, ~r,s \in [K].
    \end{align*}
\end{itemize}
\vspace{-12pt}
\end{tcolorbox}

Firstly, we need the following Lemma for the Hadamard product of matrices:
\begin{lem}\cite{horn2012matrix}\label{ppt:hadamard_tr}
For column vectors $\vx$ and $\vy$, and corresponding diagonal matrices $\mD_{\vx}$ and $\mD_{\vy}$ with these vectors as their main diagonals, the following identity holds:
\[
{\vx}^*({\mA} \circ {\mB}) {\vy} = \mathrm{tr}\left(\mD_{\vx}^* {\mA} \mD_{\vy} {\mB}^\top\right),
\]
where $\vx^*$ denotes the conjugate transpose of $\vx$.
\end{lem}

The following proof focuses on the second and third-order consensuses.
It is worth noting that, although the first-order consensus is not necessary for the derivation of the unique solution, it still helps improve the stability of solving for $\mT$ and $\vp$ numerically.

\paragraph{Step I: Transform the second-order equations.}

Denoted by $\mT_r = \mT \mS_r$, where $\mS_r$ permutes particular columns of $\mT$.
Let $\ve_i$ be the column vector with only the $i$-th element being $1$ and $0$ otherwise. 
With Lemma~\ref{ppt:hadamard_tr}, the second-order consensus can be transformed as
\[
 \ve_i^\top \vc^{[2]}_{r}  = \ve_i^\top (\mT \circ \mT_r)^\top \vp = \mathrm{tr}\left(\mD_{\ve_i} {\mT}^\top \mD_{\vp}  {\mT} \mS_r\right)
\]
Then the $(i,(i+r)_K)$-th element of matrix $ {\mT}^\top \mD_{\vp} {\mT}$ is
\[
     ({\mT}^\top \mD_{\vp} {\mT})[{i,(i+r)_K}] = \ve_i^\top \vc^{[2]}_{r}.
\]
With a fixed $\ve_i^\top \vc^{[2]}_{r}, \forall i,r \in [K]$, denote by
\begin{equation}\label{eq:prove_o2}
     {\mT}^\top \mD_{\vp} {\mT}= \mT_{\dag},
\end{equation}
where $\mT_{\dag}[{i,(i+r)_K}] = \ve_i^\top \vc^{[2]}_{r}$.
Note $\mT_\dag$ is fixed given $\vc^{[2]}_r, \forall r\in[K]$.

\paragraph{Step II: Transform the third-order equations.}
Following the idea in Step I, we can also transform the third-order equations.
First, notice that
\[
 \ve_i^\top \vc^{[3]}_{r,s}  = \ve_i^\top [  (\mT \circ \mT_s) \circ  \mT_r ]^\top \vp = \mathrm{tr}\left(\mD_{\ve_i}  (\mT \circ \mT_s)^\top \mD_{\vp} {\mT} \mS_r      \right).
\]
Then the $(i,(i+r)_K)$-th element of matrix $(\mT \circ \mT_s)^\top \mD_{\vp}  {\mT}$ is 
\[
((\mT \circ \mT_s)^\top \mD_{\vp}  {\mT})[{i,(i+r)_K}] =  \ve_i^\top \vc^{[3]}_{r,s}.
\]
With a fixed $\ve_i^\top \vc^{[3]}_{r,s}, \forall i,r \in [K]$, denote by
\begin{equation}\label{eq:prove_o31}
   (\mT \circ \mT_s)^\top \mD_{\vp}  {\mT} = \mT_{\ddag,s} \Rightarrow  {\mT}^\top \mD_{\vp}   (\mT \circ \mT_s) = \mT_{\ddag,s}^\top,
\end{equation}
where $\mT_{\ddag,s}[{i,(i+r)_K}] = \ve_i^\top \vc^{[3]}_{r,s}$.
According to Eqn.~(\ref{eq:prove_o2}), we have
\[
  {\mT}^\top \mD_{\vp}  (\mT \circ \mT_s)
  = {\mT}^\top \mD_{\vp} \mT \mT^{-1}   (\mT \circ \mT_s) 
  = \mT_{\dag} \mT^{-1} (\mT \circ \mT_s)
  = \mT_{\ddag,s}^\top.
\]
Thus
\begin{equation}\label{eq:hard_problem}
      (\mT \circ \mT_s) 
  = \mT\mT_{\dag}^{-1}\mT_{\ddag,s}^\top, \forall s\in[K].
\end{equation}

\paragraph{Step III: From matrices to vectors}

With Step I and Step II, we could transform the equations formulated by the second and the third-order consensuses to a particular system of multivariate quadratic equations of $\mT$ in Eqn.~(\ref{eq:hard_problem}).
Generally, these equations could have up to $2^{K^2}$ solutions introduced by different combinations of each element in $\mT$.
To prove the uniqueness of $\mT$, we need to exploit the structure of the equations in (\ref{eq:hard_problem}).

For a clear representation of the structure of equations and solutions, we first consider one subset of the equations in (\ref{eq:hard_problem}). Specifically, let $s=0$ we have
\begin{equation}\label{eq:eq2solve}
      (\mT \circ \mT) 
  = \mT\mT_{\dag}^{-1}\mT_{\ddag}^\top.
\end{equation}
Then we need to study the number of feasible $\mT$ satisfying Eqn.~(\ref{eq:eq2solve}).
Denote by $\mA = \mT_{\ddag}(\mT_{\dag}^{-1})^\top$.
Then each row of $\mT$, denoted by $\vu^\top$, is a solution to the equation
\begin{equation}\label{eq:vector_problem}
    \mA\vu = \mD_{\vu} \vu~~~~~\text{(a.k.a.~} \mA\vu = \vu \circ  \vu \text{)}.
\end{equation}
Till now, in Step III, we split the matrix $\mT$ to several vectors $\vu$, and transform our target from finding a matrix solution $\mT$ for (\ref{eq:hard_problem}) to a set of vector solutions $\vu$ for (\ref{eq:vector_problem}). 

Assume there are $M$ feasible $\vu$ vectors.
We collect all the possible $\vu$ and define $\mU := [\vu_1, \vu_2, \cdots, \vu_M], \vu_{i} \ne \vu_{i'}, \forall i,i' \in [M]$.
If $M=K$, we know there exists at most $K!$ different $\mT$ (considering all the possible permutations of $\vu$) that Eqn.~(\ref{eq:eq2solve}) holds.
Further, by considering an informative $\mT$ as Assumption~\ref{ap:informative}, we can identify a particular permutation.
Therefore, if $M=K$ and $\mT$ is informative, we know there exists and only exists one unique $\mT$ that Eqn.~(\ref{eq:eq2solve}) holds.

\paragraph{Step IV: Constructing the $M$-th vector}

Supposing $M>K$, we have
\[
\mA \mU = 
\mA [\vu_1, \vu_2, \cdots, \vu_K, \cdots \vu_M] = [\mD_{\vu_1}\vu_1, \mD_{\vu_2}\vu_2, \cdots, \mD_{\vu_K}\vu_K, \cdots \mD_{\vu_M}\vu_M].
\]
With a non-singular $\mT$ (Assumption~\ref{ap:invT}), without loss of generality, we will assume the first $K$ columns are full-rank.
Then $\vu_M$ must be a linear combination of the first $K$ columns, i.e., $\vu_M = \sum_{i\in[K]}\lambda_i \vu_i = \mU \vlmd_0$, where $\vlmd_0 = [\lambda_1,\lambda_2,\cdots,\lambda_K,0,\cdots,0]$.
According to the equation $\mA\vu = \mD_{\vu} \vu = \vu \circ \vu$, we have
\[
\mA\vu_M = \mD_{\vu_M} \vu_M = \mD_{\mU\vlmd_0} \mU \vlmd_0,
\]
and
\[
\mA\vu_M =  \sum_{i\in[M]} \vlmd_0[i] \mA\vu_i = \sum_{i\in[M]} \vlmd_0[i] \vu_i \circ \vu_i  = (\mU\circ \mU) \vlmd_0.
\]
Thus
\[
(\mU\circ \mU) \vlmd_0 = \mD_{\mU\vlmd_0} \mU \vlmd_0 = (\mU \vlmd_0) \circ (\mU \vlmd_0).
\]
Note that, the matrix $\mU$ can be written as $\mU =[\mU_K, \mU_{M-K}]$, and the vector $\vlmd_0$ can be written as $\vlmd_0 = [ \vlmd^\top, 0, \cdots, 0]^\top$, where $\vlmd:=[\lambda_1,\cdots,\lambda_K]^\top$.
Then the above equation can be transformed as follows:
\[
(\mU_K \circ \mU_K)\vlmd = \vu_M \circ \vu_M, \text{~and~~} \mU_K\vlmd = \vu_M.
\]
Similarly, $\forall s\in[K]$, we have
\[
(\mU_K \circ (\bar\mS_s\mU_K))\vlmd = \vu_M \circ (\bar\mS_s\vu_M), \text{~and~~} \mU_K\vlmd = \vu_M,
\]
where $\bar\mS_s \vu_M$ denotes a row circular shift such that $(\bar\mS_s \vu_M)[{i}] = \vu_M[{i+s}]$. Note $\bar\mS_s = \mS_s^\top$. 
Applying Lemma~\ref{ppt:hadamard_tr}, we have
\[
\mathrm{tr}(  \mD_{\ve_i} \mU_K \mD_{\vlmd} \mU_K^\top \bar\mS_s^\top )
=
\mathrm{tr}(  \mD_{\ve_i} \mU_K \mD_{\vlmd} \mU_K^\top \mS_s )
=
(\vu_M \circ (\bar\mS_s \vu_M))[i]
\]
Then the $(i,(i+s)_K)$-th element of matrix $\mU_K \mD_{\vlmd} \mU_K^\top$ is 
\[
(\mU_K \mD_{\vlmd} \mU_K^\top)[{i,(i+s)_K}] = (\vu_M \circ (\bar\mS_s \vu_M))[i] = \vu_M[i] \cdot \vu_M[{(i+s)_K}].
\]
Then we have
\[
\mU_K \mD_{\vlmd} \mU_K^\top = \mQ, \text{~and~~} \mQ = \vu_M \vu_M^\top.
\]
When $\mT$ is non-singular, we know $\mU$ is invertible (full-rank), then
\[
\mD_{\vlmd} = (\mU_K^{-1} \vu_M) (\mU_K^{-1} \vu_M)^\top.
\]
Thus ${\sf Rank}(\mD_{\vlmd})=1$. Recalling $\1^\top\vlmd = 1$, the vector $\vlmd$ could only be one-hot vectors, i.e. $\ve_i, \forall i\in[K]$.
This proves $\vu_M$ must be the same as one of $\vu_i, i\in[K]$.

\paragraph{Wrapping-up: Unique $\mT$}
From Step III, we know that, if $M=K$, we have a unique $\mT$ under the assumption that $\mT$ is informative and non-singular.
Step IV proves the $M$-th ($M>K$) vector $\vu$ must be identical to one of $\vu_i, i \in [K]$, indicating we only have $M=K$ non-repetitive $\vu$ vectors.
Therefore, our consensus equations are sufficient for guaranteeing a unique $\mT$.
Besides, note there is no approximation applied during the whole proof. Thus with a perfect knowledge of $\vc^{[\nu]},\nu=1,2,3,$ the unique $\mT$ satisfying the consensus equations is indeed the true noise transition matrix.

\subsection{Feasibility of Assumption $|E_3^*| = \Theta(N)$}\label{supp:feasibleE}

\begin{figure}[h]
    \centering
    \includegraphics[width = 0.25\textwidth]{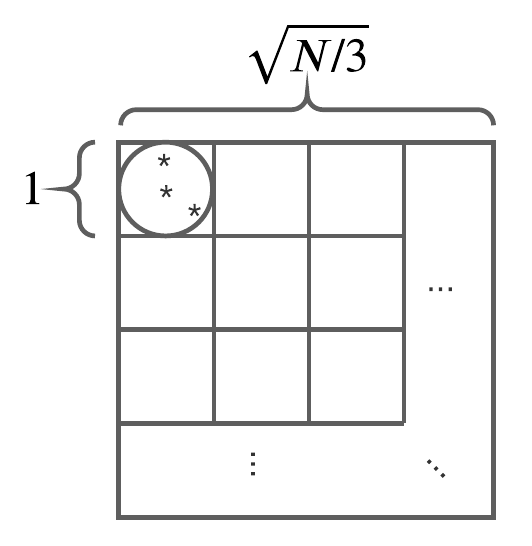}
    \caption{Illustration of a special case.}
    \label{fig:circle}
\end{figure}

We discuss the feasibility of our assumption on the number of $3$-tuples.
According to the definition of $E_3^*$, we know there are no more than $|E_3^*| \le \lfloor N/3 \rfloor$ feasible $3$-tuples. Strictly deriving the lower bound for $|E_3^*|$ is challenging due to the unknown distributions of representations. 
To roughly estimate the order of $|E_3^*|$ (i.e., the maximum number of non-overlapping $3$-tuples), we consider a special scenario where those high-dimensional representations could be mapped to a $2$-D square of width $\sqrt{N/3}$, each grid of width $1$ has exactly $3$ mapped representations, and one mapped representation is at the center of each grid (also the center of each circle). 
Consider a particular construction of feasible $3$-tuples as illustrated in Figure~\ref{fig:circle}.
We require that, for each grid, the $2$-NN fall in the corresponding circle.
Otherwise, they may become the $2$-NN of representations in other nearby girds.
Assume the $2$-NN are independently and uniformly distributed in the unit square, thus the probability of both $2$-NN falling in the circle is $(\pi/4)^2$. Noting there are $N/3$ grids in the big square illustrated in Figure~\ref{fig:circle}, the expected number of feasible $3$-tuples in this case is $\frac{\pi^2}{48}\cdot N = \Theta(N)$.
Although this example only considers a special case, it demonstrates the order of $|E_3^*|$ could be $\Theta(N)$ with appropriate representations.

\subsection{Proof for Lemma~\ref{lem:sample_c}}\label{supp:proof_c}

Then we present the proof for Lemma~\ref{lem:sample_c}.
\begin{proof}
Recall in Eqn.~(\ref{eq:EM_HOC}), each high-order consensus pattern could be estimated by the sample mean of $|E_3^*|$ independent and identically distributed random variables, thus according to Hoeffding's inequality \cite{hoeffding1963probability}, w.p. $1-\delta$, we have
\[
|\hat \vc^{[i]}[j] - \vc^{[i]}[j]| \le  \sqrt{\frac{\ln \frac{2}{\delta}}{2|E_3^*|}}, i=1,2,3, \forall j,
\]
which is at the order of $O(\sqrt{{\ln(1/\delta)}/{N}})$.
\end{proof}

\subsection{Proof for Theorem~\ref{thm:sample}}\label{supp:proof_sample}

Consider a particular uniform off-diagonal matrix $\mT$, where the off-diagonal elements are $T_{ij} = \frac{1-T_{ii}}{K-1}$.
Recall the clean prior probability for the $i$-th class is $p_i$.
To find the upper bound for the sample complexity, we can only consider a subset of our consensus equations.
Specifically, we consider the equations related to the $i$-th element of Eqn.~(\ref{eq:o1}) and Eqn.~(\ref{eq:o2}) when $r=0$.
Then a solution to our consensus equations will need to satisfy at least the following two equations:
\begin{align}
    \hat{p}_i \hat{T}_{ii} + (1-\hat{p}_i) \frac{1-\hat{T}_{ii}}{K-1} = \hat{c}_1, \label{eqn:1}\\
    \hat{p}_i \hat{T}^2_{ii} + (1-\hat{p}_i) \frac{(1-\hat{T}_{ii})^2}{(K-1)^2} = \hat{c}_{2}, \label{eqn:2}
\end{align}
where $\hat{p}_i$ and $\hat{T}_{ii}$ denote the estimated clean prior probability and noisy transition matrix, $\hat{c}_1$ and $\hat{c}_{2}$ denote the corresponding estimates of first- and second-order statistics. Lemma~\ref{lem:sample_c} shows, with probability $1-\delta$:
 \[
 |\hat{c}_i - c_i| \leq O\left(\sqrt{\frac{\ln(1/\delta)}{N}}\right).
 \]
 Multiplying both sides of Eqn. (\ref{eqn:1}) by $T_{ii}$ and adding Eqn. (\ref{eqn:2}), we have
 \[
 K(K-1)\hat{p}_i \hat{T}^2_{ii} + (1-\hat{p}_i)(1-\hat{T}_{ii}) = (K-1)\hat{c}_1 \hat{T}_{ii} +  (K-1)^2\hat{c}_2.
 \]
 Note the above equality also holds for the true values $p_i,T_{ii},c_1,c_2$. Taking the difference we have 
  \begin{align*}
  &(\hat{T}_{ii}-T_{ii})(K(K-1)p_i (T_{ii}+\hat{T}_{ii}) - (1-p_i)-(K-1)c_1)  
  \\
  =& (K-1)^2 (\hat{c}_2-c_2) + (K-1)(\hat{c}_1 - c_1) \hat{T}_{ii} - K(K-1)\hat T_{ii}^2 (\hat p_i - p_i) - (\hat T_{ii}-1) (\hat p_i - p_i).
 \end{align*}
 Taking the absolute value for both sides yields
  \begin{align*}
  & |\hat{T}_{ii}-T_{ii}| \cdot |K(K-1)p_i (T_{ii}+\hat{T}_{ii}) - (1-p_i)-(K-1)c_1|  \\
  \leq &  (K-1)^2 |\hat{c}_2-c_2|  + (K-1)|\hat{c}_1 - c_1| + (K(K-1)+1)  |\hat p_i - p_i|
 \end{align*}
 From Eqn.~(\ref{eqn:1}), we have
 \[
 \hat p_i = \frac{K-1}{K} \frac{\hat c_1 - 1/K}{\hat T_{ii} - 1/K} + \frac{1}{K}.
 \]
 Thus
  \[
 |\hat p_i - p_i| \le  \frac{K-1}{K} \frac{|\hat c_1 - c_1|}{\min(\hat T_{ii}, T_{ii})-1/K},
 \]
 indicating $ |\hat p_i - p_i|$ is at the order of $|\hat c_1 - c_1|$.
Note that 
 \begin{align*}
 K(K-1)p_i (T_{ii}+\hat{T}_{ii}) - (1-p_i)-(K-1)c_1 \geq  K(K-1)p_i T_{ii} - (1-p_i)-(K-1) c_1.   
 \end{align*}
When $K(K-1)p_i T_{ii} - (1-p_i)-(K-1) c_1 > 0$, we have
 \begin{align*}
      |\hat{T}_{ii}-T_{ii}| \leq \frac{(K-1)^2 |\hat{c}_2-c_2|  + (K-1)|\hat{c}_1 - c_1| + (K(K-1)+1) \frac{K-1}{K} \frac{|\hat c_1 - c_1|}{\min(\hat T_{ii}, T_{ii})-1/K} }{K(K-1)p_i T_{ii} - (1-p_i)-(K-1) c_1}.
 \end{align*}
 Then by union bound we know, w.p. $1-2\delta$, the estimation error $|\hat{T}_{ii} - T_{ii}|$ is at the same order as $|\hat{c}_i - c_i|$, i.e. $O(\sqrt{\frac{\ln(1/\delta)}{N}})$.

\section{More Discussions}\label{supp:discuss}

\subsection{Soft $2$-NN Label Clusterability}\label{supp:relax2nn}

The soft $2$-NN label clusterability means one's $2$-NN may have a certain (but small) probability of belonging to different clean classes.
Statistically, if we use a new matrix $\mT^{\text{soft}}$ to characterize the probability of getting a different nearest neighbor, i.e. $T^{\text{soft}}_{ij} = \PP(Y_{2} = j | Y_1=i) = \PP(Y_{3} = j | Y_1=i)$, the second-order consensuses become $\vc^{[2]}_{r} := (\mT\circ (\mT^{\text{soft}}\mT_r))^\top \vp$ and the third-order consensuses become $ \vc^{[3]}_{r,s} := (\mT\circ (\mT^{\text{soft}}\mT_r) \circ (\mT^{\text{soft}}\mT_s))^\top  \vp$. Specifically, if $T^{\text{soft}}_{ij} = e, \forall i\ne j$ and $T^{\text{soft}}_{ii} = 1-(K-1)e, 0 \le e<1/K$, where $e$ captures the small perturbation of the 2-NN assumption, our solution will likely output a transition matrix that affects the label noise between the effects of $\mT^{\text{soft}}\mT$ and $\mT$. The above observation informs us that our estimation will be away from the true $\mT$ by at most a factor $e$. %
When $e=0$, we recover the original $2$-NN label clusterability condition.

\subsection{Local \texorpdfstring{$\mT(X)$}{}}\label{supp:localT}

\paragraph{Sparse regularizer}
Compared with estimating one global $\mT$ using the whole dataset of size $N$, each local estimation will have access to only $M$ instances, where $M\ll N$. Thus the feasibility of returning an accurate $\mT(x_n)$ requires more consideration. 
In some particular cases, e.g., \ours{} Local in Table~\ref{table:cifar-inst}, when $\vp$ is sparse due to the local datasets,
we usually add a regularizer to ensure a sparse $\vp$, such as $\sum_{i\in[K]}\ln (c_i + \varepsilon), \varepsilon \rightarrow 0_+$, where $c_i$ is the $i$-th element of $\vp$. Note the standard sparse regularizer, i.e. $\ell_1$-norm $\|\vp\|_1$, could not be applied here since $\|\vp\|_1 = 1$.
Therefore, with a regularizer that shrinks the search space and fewer variables, we could get an accurate estimate of $T(X)$ with a small $M$.

\paragraph{Other extensions}
Even with $M$-NN noise clusterability, estimating $\mT(X)$ for the whole dataset requires executing Algorithm~\ref{alg:Test} a numerous number of times ($\sim N/M$).
If equipped with prior knowledge that the label noise can be divided into several groups and $\mT=\mT(X)$ within each group \cite{xia2020parts,wang2020fair}, we only need to estimate $\mT$ for each group by treating instances in each group as a local dataset and directly apply Algorithm~\ref{alg:Test}.
As a preliminary work on estimating $\mT$ relying on clusterability, the focus of this paper is to provide a generic method for estimating $\mT$ given a dataset. Designing efficient algorithms to split the original dataset into a tractable number of local datasets is interesting for future investigation.

\subsection{Feasibility of Assumption~\ref{ap:invT} and Assumption~\ref{ap:informative}}\label{supp:feasibilityAP}
\begin{enumerate}
    \item Denote the confusion matrix by $\mC[h]$, where each element is $C_{ij}[h]:=\PP(Y=i,h(X)=j)$ and $h(X)=j$ represents the event that the classifier predicts $j$ given feature $X$. Then the noisy confusion matrix could be written as $\widetilde\mC[h]:= \mT^\top \mC[h]$.
If $\mT$ is non-singular (a.k.a. invertible), statistically, we can always find the inverse matrix $\mT^{-1}$ such that the clean confusion matrix could be recovered as $\mC[h]=(\mT^{-1})^\top \widetilde\mC[h]$. Otherwise, we may think the label noise is too ``much'' such that the clean confusion matrix is not recoverable by $\mT$. Then learning $\mT$ may not be meaningful anymore. Therefore, Assumption~\ref{ap:invT} is effectively ensuring the necessity of estimating $\mT$.
    \item We require $T_{ii} > T_{ij}$ in Assumption~\ref{ap:informative} to ensure instances from observed class $i$ (observed from noisy labels) are informative \cite{liu2017machine}.
Intuitively, this assumption characterizes a particular permutation of row vectors in $\mT$. Otherwise, there may exist $K!$ possible solutions by considering all the permutations of $K$ rows \cite{liu2020surrogate}. 
\end{enumerate}

\section{More Detailed Experiment Settings}\label{supp:exp_set}

\subsection{Generating the Instance-Dependent Label Noise }\label{sec:instance_noise_gen}
In this section, we introduce how to generate instance-based label noise, which is illustrated in Algorithm \ref{algorithm1}.
Note this algorithm follows the state-of-the-art method \cite{xia2020parts,zhu2020second}.
Define the noise rate (the global flipping rate) as $\eta$.
To calculate the probability of $x_{n}$ mapping to each class under certain noise conditions, we set sample instance flip rates $q_{n}$ and sample parameters $W$. The size of $W$ is $ S\times K $, where $S$ denotes the length of each feature.

First, we sample instance flip rates $q_{n}$ from a truncated normal distribution $\mathbf{N}(\eta, 0.1^{2}, [0, 1])$ in Line~2. The average flipping rate (a.k.a. average noise rate) is $\eta$. $q_n$ avoids all the instances having the same flip rate. Then, in Line~3, we sample parameters $W$ from the standard normal distribution for generating the instance-dependent label noise. Each column of $W$ acts as a projection vector. After acquiring $q_{n}$ and $W$, we can calculate the probability of getting a wrong label for each instance$(x_{n},y_n)$ in Lines~4~--~6. 
Note that in Line~5, we set $p_{y_{n}} = -\infty$, which ensures that $x_{n}$ will not be mapped to its own true label. In addition, Line~7 ensures the sum of all the entries of $p$ is 1.
Suppose there are two features: ${x}_{i}$ and ${x}_{j}$ where ${x}_{i} = {x}_{j}$. Then the possibility $p$ of these two features, calculated by ${x}\cdot W$, from the Algorithm~\ref{algorithm1}, would be exactly the same. Thus the label noise is strongly instance-dependent.

Note Algorithm~\ref{algorithm1} cannot ensure $T_{ii}(X)>T_{ij}(X)$ when $\eta>0.5$. To generate an informative dataset, we set $0.9 \cdot T_{ii}(X)$ as the upper bound of $T_{ij}(X)$ and distribute the remaining probability to other classes.

	\begin{algorithm*}[!t]
		\caption{Instance-Dependent Label Noise Generation}
		\label{algorithm1}
		\begin{algorithmic}[1]
			\renewcommand{\algorithmicrequire}{\textbf{Input:}}
			\renewcommand{\algorithmicensure}{\textbf{Iteration:}}
			\REQUIRE ~~\\
			 1: Clean examples ${({x}_{n},y_{n})}_{n=1}^{N}$; Noise rate: $\eta$; Size of feature: $1\times S$; Number of classes: $K$.
			\ENSURE ~~\\
			2: Sample instance flip rates $q_n$ from the truncated normal distribution $\mathcal{N}(\eta, 0.1^{2}, [0, 1])$;\\
            3: Sample   $W \in \mathcal{R}^{S \times K}$ from the standard normal distribution $\mathcal{N}(0,1^{2})$;\\
			\textbf{for} $n = 1$ to $N$  \textbf{do} \\
			4: \qquad $p = {x}_{n} \cdot  W$   ~~~~ \algcom{//  Generate instance dependent flip rates. The size of $p$ is $1\times K$.} 
			\\
			5: \qquad  $p_{y_{n}} = -\infty$   ~~~~~\algcom{//  Only consider entries that are different from the true label}\\
			6: \qquad $p = q_{n} \cdot \texttt{SoftMax}(p) $ ~~~~ \algcom{//  Let $q_n$ be the probability of getting a wrong label}
			\\
			7: \qquad $p_{y_{n}} = 1 - q_{n}$ ~~~~\algcom{//  Keep clean w.p. $1 - q_{n}$} 
			\\
			8: \qquad Randomly choose a label from the label space as noisy label $\tilde{y}_{n}$ according to $p$;
			\\
			\textbf{end for}\\
			
			\renewcommand{\algorithmicensure}{\textbf{Output:}}
			\ENSURE ~~\\
			9: Noisy examples ${({x}_{i},\tilde{y}_{n})}_{n=1}^{N}$.
		\end{algorithmic}
	\end{algorithm*}
	
\subsection{Basic Hyper-Parameters}\label{supp:hyper}

To testify the classification performance, we adopt the flow: 1) Pre-training $\rightarrow$ 2) Global Training $\rightarrow$ 3) Local Training.
Our \ours{} estimator is applied once at the beginning of each above step.
Each training stage re-trains the model.
In Stage-1, we load the standard ResNet50 model pre-trained on ImageNet to obtain basic representations.
At the beginning of Stage-2 and Stage-3, we use the representations given by the current model.
All experiments are repeated three times.
\emph{\ours{} Global} only employs one global $\mT$ with $G=50$ and $|E|=15k$ as inputs of Algorithm~\ref{algorithm1}. \emph{\ours{} Local} uses $300$ local matrices %
($250$-NN noise clusterability, $|D_{h(n)}|=250$, $G=30$, $|E|=100$) for CIFAR-10 and $5$ local matrices ($10k$-NN noise clusterability, $|D_{h(n)}|=10k$, $G=30$, $|E|=5k$) for CIFAR-100.
Note the local matrices may not cover the whole dataset. For those uncovered instances, we simply apply $\mT$.

\paragraph{Other hyperparameters:}
\squishlist 
    \item Batch size: 128 (CIFAR), 32 (Clothing1M)
    \item Learning rate: 
    \begin{itemize}
        \item CIFAR-10: Pre-training: $0.1$ for $20$ epochs $\rightarrow$ $0.01$ for $20$ epochs. Global Training: $0.1$ for $20$ epochs $\rightarrow$ $0.01$ for $20$ epochs. Local Training: $0.1$ for $60$ epochs $\rightarrow$ $0.01$ for $60$ epochs $\rightarrow$ $0.001$ for $60$ epochs.
        \item CIFAR-100: Pre-training: $0.1$ for $30$ epochs $\rightarrow$ $0.01$ for $30$ epochs. Global Training: $0.1$ for $30$ epochs $\rightarrow$ $0.01$ for $30$ epochs. Local Training: $0.1$ for $30$ epochs $\rightarrow$ $0.01$ for $30$ epochs $\rightarrow$ $0.001$ for $30$ epochs.
        \item Clothing1M: $0.01$ for $25$ epochs $\rightarrow$ $0.001$ for $25$ epochs $\rightarrow$ $0.0001$ for $15$ epochs $\rightarrow$ $0.00001$ for $15$ epochs (Pre-training, Global training, and local training)
    \end{itemize}
\item Momentum: 0.9
\item Weight decay: 0.0005 (CIFAR) and 0.001 (Clothing1M)
\item Optimizer: SGD (Model training) and Adam with initial a learning rate of $0.1$ (solving for $\mT$)
\squishend

For each epoch in Clothing1M, we sample 1000 mini-batches from the training data while ensuring the (noisy) labels are balanced. 
The global $\mT$ is obtained by an average of $\mT$ from $5$ random epochs.
We only use $\mT(X) = \mT$ in local training.
Estimating local transition matrices using \ours{} on Clothing1M is feasible, e.g., assuming $M$-NN noise clusterability, but it may be time-consuming to tune $M$.
Noting our current performance is already satisfying, and the focus of this paper is on the ability to estimate $\mT$, we leave the combination of $\mT(X)$ with loss correction or other advanced techniques for future works.

\subsection{Global and Local Estimation Errors on CIFAR-10 with Human Noise}\label{supp:est_error_local}

Algorithm~\ref{algorithm2} details the generation of local datasets. 
Notice the fact that the $i$-th row of $\mT(x_n)$ could be any feasible values when $p_i = 0$, so as the estimates $\hat\mT_{\textsf{local}}$.
In such case, we need to refer to $\mT$ to complete the information.
Particularly, we calculate the weighted average value with the corresponding $\hat\mT$ as
\begin{equation*}
    \hat\mT_{\textsf{local}}[i] = (1-\zeta + \hat p_i) \hat\mT_{\textsf{local}}[i] + (\zeta- \hat p_i) \hat\mT[i],
\end{equation*}
where $\hat\mT_{\textsf{local}}[i]$ and $\hat\mT[i]$ denote the $i$-th row of estimates $\hat\mT_{\textsf{local}}$ and $\hat\mT$, $\hat p_i$ denotes the estimated clean prior probability of class-$i$ given the local dataset.
We use $\zeta=1$ for local estimates of CIFAR-10, and $\zeta=0.5$ for local estimate of CIFAR-100.

\begin{figure}
    \centering
    \includegraphics[width=0.45\textwidth]{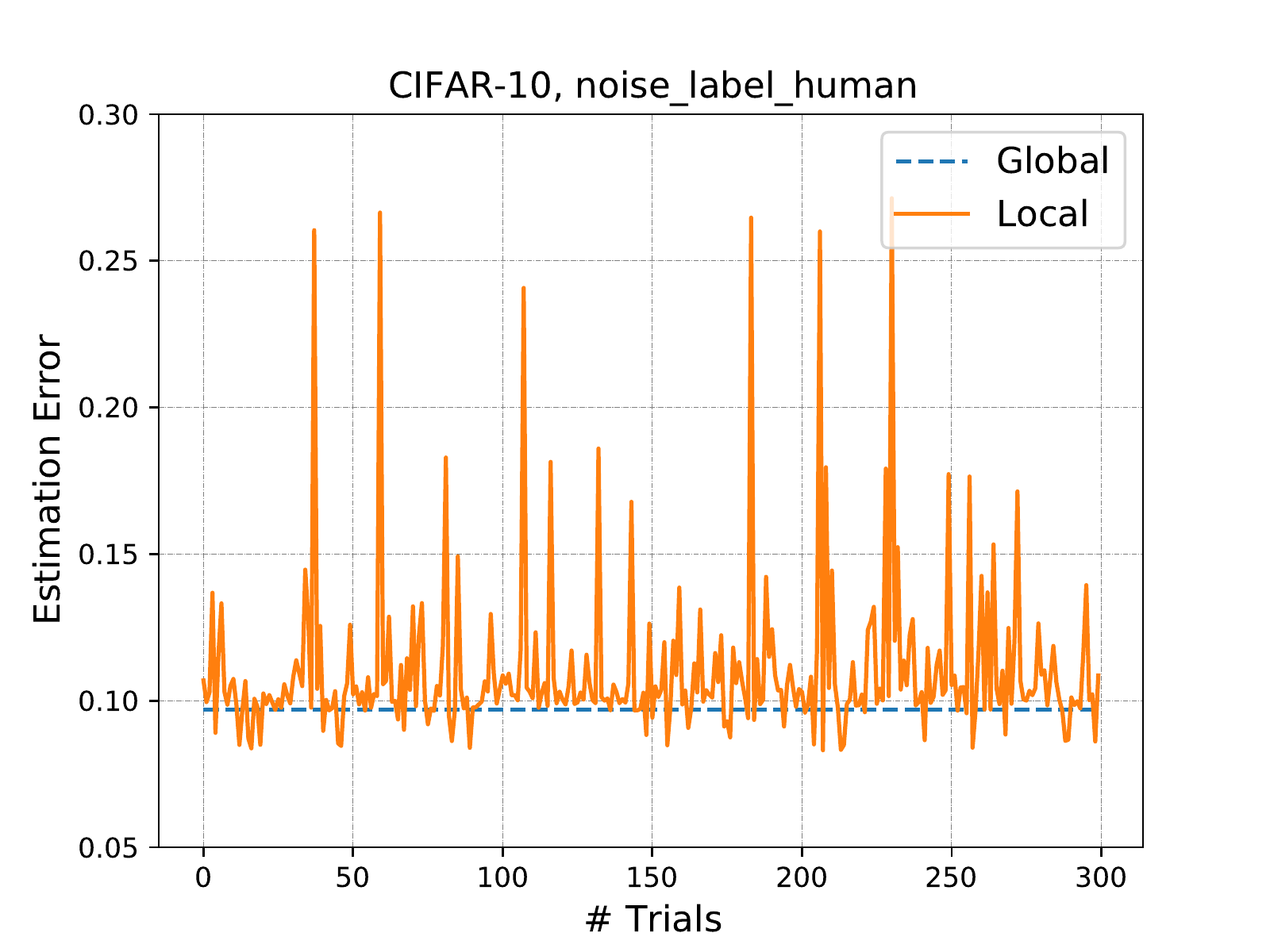}
    \caption{{Illustration of the global and local estimation errors. Global estimation error: 0.0970. Local estimation errors: mean = 0.1103, standard deviation = 0.0278.}}
    \label{fig:local_illu}
\end{figure}

Figure~\ref{fig:local_illu} illustrates the variation of local estimation errors on CIFAR-10 with human noise using \ours{}.
	\begin{algorithm*}[!t]
		\caption{Local Datasets Generation}
		\label{algorithm2}
		\begin{algorithmic}[1]
			\renewcommand{\algorithmicrequire}{\textbf{Input:}}
			\renewcommand{\algorithmicensure}{\textbf{Iteration:}}
			\REQUIRE ~~\\
			 1: 
			 Maximal rounds: $G'$. Local dataset size: $L$. Noisy dataset: $\widetilde{D}=\{( x_n, \tilde y_n)\}_{n\in  [N]}$. Noisy dataset size: $|D|$. 
			\ENSURE ~~\\
			2:
			Initialize the $|D|$-dimensional index list: $S=\1$\\
			\textbf{for} $k = 1$ to $G'$  \textbf{do} \\
            \qquad\textbf{if}({${\sf size}(S[S>0])>0$}) \textbf{then}\\
            3:\qquad$\text{Idx}_{\sf selected} = \texttt{random.choice}(S[S>0])$ ~~~~ 
			\algcom{// Choose a local center index randomly from the unselected index of $\widetilde{D}$.}\\
			\qquad\textbf{else}\\
			4:\qquad$\text{Idx}_{\sf selected} = \texttt{random.randint}(0, |D|)$  ~~~~ 
			\algcom{// If the selected index has covered $\widetilde{D}$, we choose local center randomly.}\\
			\qquad\textbf{end  if}\\
			5:\qquad$\text{Idx}_{\sf local} = \texttt{ SelectbyDist}(\text{Idx}_{\sf selected}, L)$ ~~~~ 
			\algcom{// Select the index of $L$ features closest to $\text{Idx}_{\sf selected}$.}\\
			6:\qquad$S[\text{Idx}_{\sf local}] = -1$ ~~~~ 
			\algcom{// Mark the state of the selected index in $S$ to avoid duplicate selection.}\\
			7:\qquad$\widetilde D_k = \widetilde{D}[\text{Idx}_{\sf local}]$ ~~~~ 
			\algcom{// Build a local dataset by selecting $( x_i, \tilde y_i), i \in \text{Idx}_{local}$.}\\
			\textbf{end for}\\
			\renewcommand{\algorithmicensure}{\textbf{Output:}}
			\ENSURE ~~\\
			8: Local Datasets
			$\widetilde D_k = \{(x_n,\tilde{y}_n)\} \cup \{(x_{n_1},\tilde y_{n_1}), \cdots, (x_{n_M},\tilde y_{n_M})\}$, $n_i, k \in [L], i\in [M]$.
		\end{algorithmic}
	\end{algorithm*}

%% file: main.bbl
\begin{thebibliography}{53}
\providecommand{\natexlab}[1]{#1}
\providecommand{\url}[1]{\texttt{#1}}
\expandafter\ifx\csname urlstyle\endcsname\relax
  \providecommand{\doi}[1]{doi: #1}\else
  \providecommand{\doi}{doi: \begingroup \urlstyle{rm}\Url}\fi

\bibitem[Amid et~al.(2019{\natexlab{a}})Amid, Warmuth, Anil, and
  Koren]{amid2019robust}
Amid, E., Warmuth, M.~K., Anil, R., and Koren, T.
\newblock Robust bi-tempered logistic loss based on bregman divergences.
\newblock In \emph{Advances in Neural Information Processing Systems}, pp.\
  14987--14996, 2019{\natexlab{a}}.

\bibitem[Amid et~al.(2019{\natexlab{b}})Amid, Warmuth, and
  Srinivasan]{amid2019two}
Amid, E., Warmuth, M.~K., and Srinivasan, S.
\newblock Two-temperature logistic regression based on the tsallis divergence.
\newblock In \emph{The 22nd International Conference on Artificial Intelligence
  and Statistics}, pp.\  2388--2396. PMLR, 2019{\natexlab{b}}.

\bibitem[Bengio et~al.(2013)Bengio, Courville, and
  Vincent]{bengio2013representation}
Bengio, Y., Courville, A., and Vincent, P.
\newblock Representation learning: A review and new perspectives.
\newblock \emph{IEEE transactions on pattern analysis and machine
  intelligence}, 35\penalty0 (8):\penalty0 1798--1828, 2013.

\bibitem[Berthon et~al.(2021)Berthon, Han, Niu, Liu, and
  Sugiyama]{berthon2020confidence}
Berthon, A., Han, B., Niu, G., Liu, T., and Sugiyama, M.
\newblock Confidence scores make instance-dependent label-noise learning
  possible.
\newblock In \emph{Proceedings of the 38th International Conference on Machine
  Learning}, ICML, 2021.

\bibitem[Boyd et~al.(2004)Boyd, Boyd, and Vandenberghe]{boyd2004convex}
Boyd, S., Boyd, S.~P., and Vandenberghe, L.
\newblock \emph{Convex optimization}.
\newblock Cambridge university press, 2004.

\bibitem[Cheng et~al.(2020)Cheng, Zhu, Li, Gong, Sun, and Liu]{sieve2020}
Cheng, H., Zhu, Z., Li, X., Gong, Y., Sun, X., and Liu, Y.
\newblock Learning with instance-dependent label noise: A sample sieve
  approach, 2020.

\bibitem[Feldman(2020)]{feldman2020does}
Feldman, V.
\newblock Does learning require memorization? a short tale about a long tail.
\newblock In \emph{Proceedings of the 52nd Annual ACM SIGACT Symposium on
  Theory of Computing}, pp.\  954--959, 2020.

\bibitem[Gao et~al.(2016)Gao, Yang, and Zhou]{gao2016resistance}
Gao, W., Yang, B.-B., and Zhou, Z.-H.
\newblock On the resistance of nearest neighbor to random noisy labels.
\newblock \emph{arXiv preprint arXiv:1607.07526}, 2016.

\bibitem[Ghosh et~al.(2017)Ghosh, Kumar, and Sastry]{ghosh2017robust}
Ghosh, A., Kumar, H., and Sastry, P.
\newblock Robust loss functions under label noise for deep neural networks.
\newblock In \emph{Thirty-First AAAI Conference on Artificial Intelligence},
  2017.

\bibitem[Gong et~al.(2018)Gong, Li, Meng, Miao, and Liu]{gong2018decomposition}
Gong, M., Li, H., Meng, D., Miao, Q., and Liu, J.
\newblock Decomposition-based evolutionary multiobjective optimization to
  self-paced learning.
\newblock \emph{IEEE Transactions on Evolutionary Computation}, 23\penalty0
  (2):\penalty0 288--302, 2018.

\bibitem[Han et~al.(2018)Han, Yao, Yu, Niu, Xu, Hu, Tsang, and
  Sugiyama]{han2018co}
Han, B., Yao, Q., Yu, X., Niu, G., Xu, M., Hu, W., Tsang, I., and Sugiyama, M.
\newblock Co-teaching: Robust training of deep neural networks with extremely
  noisy labels.
\newblock In \emph{Advances in neural information processing systems}, pp.\
  8527--8537, 2018.

\bibitem[Han et~al.(2020)Han, Yao, Liu, Niu, Tsang, Kwok, and
  Sugiyama]{han2020survey}
Han, B., Yao, Q., Liu, T., Niu, G., Tsang, I.~W., Kwok, J.~T., and Sugiyama, M.
\newblock A survey of label-noise representation learning: Past, present and
  future.
\newblock \emph{arXiv preprint arXiv:2011.04406}, 2020.

\bibitem[Han et~al.(2019)Han, Luo, and Wang]{han2019deep}
Han, J., Luo, P., and Wang, X.
\newblock Deep self-learning from noisy labels.
\newblock In \emph{Proceedings of the IEEE International Conference on Computer
  Vision}, pp.\  5138--5147, 2019.

\bibitem[He et~al.(2016)He, Zhang, Ren, and Sun]{he2016deep}
He, K., Zhang, X., Ren, S., and Sun, J.
\newblock Deep residual learning for image recognition.
\newblock In \emph{Proceedings of the IEEE conference on computer vision and
  pattern recognition}, pp.\  770--778, 2016.

\bibitem[Hoeffding(1963)]{hoeffding1963probability}
Hoeffding, W.
\newblock Probability inequalities for sums of bounded random variables.
\newblock \emph{Journal of the American Statistical Association}, 58\penalty0
  (301):\penalty0 13--30, 1963.
\newblock ISSN 01621459.

\bibitem[Horn \& Johnson(2012)Horn and Johnson]{horn2012matrix}
Horn, R.~A. and Johnson, C.~R.
\newblock \emph{Matrix analysis}.
\newblock Cambridge university press, 2012.

\bibitem[Ji et~al.(2019)Ji, Henriques, and Vedaldi]{ji2019invariant}
Ji, X., Henriques, J.~F., and Vedaldi, A.
\newblock Invariant information clustering for unsupervised image
  classification and segmentation.
\newblock In \emph{Proceedings of the IEEE/CVF International Conference on
  Computer Vision}, pp.\  9865--9874, 2019.

\bibitem[Jiang et~al.(2018)Jiang, Zhou, Leung, Li, and
  Fei-Fei]{jiang2017mentornet}
Jiang, L., Zhou, Z., Leung, T., Li, L.-J., and Fei-Fei, L.
\newblock Mentornet: Learning data-driven curriculum for very deep neural
  networks on corrupted labels.
\newblock In \emph{International Conference on Machine Learning}, pp.\
  2304--2313. PMLR, 2018.

\bibitem[Kolesnikov et~al.(2019)Kolesnikov, Zhai, and
  Beyer]{kolesnikov2019revisiting}
Kolesnikov, A., Zhai, X., and Beyer, L.
\newblock Revisiting self-supervised visual representation learning.
\newblock In \emph{Proceedings of the IEEE/CVF Conference on Computer Vision
  and Pattern Recognition}, pp.\  1920--1929, 2019.

\bibitem[Krizhevsky et~al.(2009)Krizhevsky, Hinton,
  et~al.]{krizhevsky2009learning}
Krizhevsky, A., Hinton, G., et~al.
\newblock Learning multiple layers of features from tiny images.
\newblock Technical report, Citeseer, 2009.

\bibitem[Li et~al.(2020)Li, Zhang, Xu, Dickerson, and Ba]{li2020noisy}
Li, J., Zhang, M., Xu, K., Dickerson, J.~P., and Ba, J.
\newblock Noisy labels can induce good representations.
\newblock \emph{arXiv preprint arXiv:2012.12896}, 2020.

\bibitem[Li et~al.(2021)Li, Liu, Han, Niu, and Sugiyama]{li2021provably}
Li, X., Liu, T., Han, B., Niu, G., and Sugiyama, M.
\newblock Provably end-to-end label-noise learning without anchor points.
\newblock \emph{arXiv preprint arXiv:2102.02400}, 2021.

\bibitem[Liu et~al.(2012)Liu, Peng, and Ihler]{liu2012variational}
Liu, Q., Peng, J., and Ihler, A.
\newblock Variational inference for crowdsourcing.
\newblock In \emph{Proceedings of the 25th International Conference on Neural
  Information Processing Systems-Volume 1}, pp.\  692--700, 2012.

\bibitem[Liu \& Tao(2015)Liu and Tao]{liu2015classification}
Liu, T. and Tao, D.
\newblock Classification with noisy labels by importance reweighting.
\newblock \emph{IEEE Transactions on pattern analysis and machine
  intelligence}, 38\penalty0 (3):\penalty0 447--461, 2015.

\bibitem[Liu(2021)]{liu2021importance}
Liu, Y.
\newblock The importance of understanding instance-level noisy labels.
\newblock In \emph{Proceedings of the 38th International Conference on Machine
  Learning}, ICML '21, 2021.

\bibitem[Liu \& Chen(2017)Liu and Chen]{liu2017machine}
Liu, Y. and Chen, Y.
\newblock Machine-learning aided peer prediction.
\newblock In \emph{Proceedings of the 2017 ACM Conference on Economics and
  Computation}, pp.\  63--80, 2017.

\bibitem[Liu \& Guo(2020)Liu and Guo]{liu2019peer}
Liu, Y. and Guo, H.
\newblock Peer loss functions: Learning from noisy labels without knowing noise
  rates.
\newblock In \emph{Proceedings of the 37th International Conference on Machine
  Learning}, ICML '20, 2020.

\bibitem[Liu \& Liu(2015)Liu and Liu]{liu2015online}
Liu, Y. and Liu, M.
\newblock An online learning approach to improving the quality of
  crowd-sourcing.
\newblock \emph{ACM SIGMETRICS Performance Evaluation Review}, 43\penalty0
  (1):\penalty0 217--230, 2015.

\bibitem[Liu et~al.(2020)Liu, Wang, and Chen]{liu2020surrogate}
Liu, Y., Wang, J., and Chen, Y.
\newblock Surrogate scoring rules.
\newblock In \emph{Proceedings of the 21st ACM Conference on Economics and
  Computation}, pp.\  853--871, 2020.

\bibitem[Lukasik et~al.(2020)Lukasik, Bhojanapalli, Menon, and
  Kumar]{lukasik2020does}
Lukasik, M., Bhojanapalli, S., Menon, A., and Kumar, S.
\newblock Does label smoothing mitigate label noise?
\newblock In \emph{International Conference on Machine Learning}, pp.\
  6448--6458. PMLR, 2020.

\bibitem[Natarajan et~al.(2013)Natarajan, Dhillon, Ravikumar, and
  Tewari]{natarajan2013learning}
Natarajan, N., Dhillon, I.~S., Ravikumar, P.~K., and Tewari, A.
\newblock Learning with noisy labels.
\newblock In \emph{Advances in neural information processing systems}, pp.\
  1196--1204, 2013.

\bibitem[Northcutt et~al.(2021)Northcutt, Jiang, and
  Chuang]{northcutt2021confident}
Northcutt, C., Jiang, L., and Chuang, I.
\newblock Confident learning: Estimating uncertainty in dataset labels.
\newblock \emph{Journal of Artificial Intelligence Research}, 70:\penalty0
  1373--1411, 2021.

\bibitem[Northcutt et~al.(2017)Northcutt, Wu, and
  Chuang]{northcutt2017learning}
Northcutt, C.~G., Wu, T., and Chuang, I.~L.
\newblock Learning with confident examples: Rank pruning for robust
  classification with noisy labels.
\newblock \emph{UAI}, 2017.

\bibitem[Patrini et~al.(2017)Patrini, Rozza, Krishna~Menon, Nock, and
  Qu]{patrini2017making}
Patrini, G., Rozza, A., Krishna~Menon, A., Nock, R., and Qu, L.
\newblock Making deep neural networks robust to label noise: A loss correction
  approach.
\newblock In \emph{Proceedings of the IEEE Conference on Computer Vision and
  Pattern Recognition}, pp.\  1944--1952, 2017.

\bibitem[Scott(2015)]{scott2015rate}
Scott, C.
\newblock A rate of convergence for mixture proportion estimation, with
  application to learning from noisy labels.
\newblock In \emph{AISTATS}, 2015.

\bibitem[Shu et~al.(2020)Shu, Zhao, Chen, Xu, and Meng]{shu2020learning}
Shu, J., Zhao, Q., Chen, K., Xu, Z., and Meng, D.
\newblock Learning adaptive loss for robust learning with noisy labels.
\newblock \emph{arXiv preprint arXiv:2002.06482}, 2020.

\bibitem[Van~Rooyen \& Williamson(2017)Van~Rooyen and
  Williamson]{van2017theory}
Van~Rooyen, B. and Williamson, R.~C.
\newblock A theory of learning with corrupted labels.
\newblock \emph{J. Mach. Learn. Res.}, 18\penalty0 (1):\penalty0 8501--8550,
  2017.

\bibitem[Wang et~al.(2021)Wang, Liu, and Levy]{wang2020fair}
Wang, J., Liu, Y., and Levy, C.
\newblock Fair classification with group-dependent label noise.
\newblock FAccT, pp.\  526–536, New York, NY, USA, 2021.

\bibitem[Wang et~al.(2019)Wang, Ma, Chen, Luo, Yi, and
  Bailey]{wang2019symmetric}
Wang, Y., Ma, X., Chen, Z., Luo, Y., Yi, J., and Bailey, J.
\newblock Symmetric cross entropy for robust learning with noisy labels.
\newblock In \emph{Proceedings of the IEEE International Conference on Computer
  Vision}, pp.\  322--330, 2019.

\bibitem[Wei et~al.(2020)Wei, Feng, Chen, and An]{wei2020combating}
Wei, H., Feng, L., Chen, X., and An, B.
\newblock Combating noisy labels by agreement: A joint training method with
  co-regularization.
\newblock In \emph{Proceedings of the IEEE/CVF Conference on Computer Vision
  and Pattern Recognition}, pp.\  13726--13735, 2020.

\bibitem[Wei \& Liu(2021)Wei and Liu]{wei2021when}
Wei, J. and Liu, Y.
\newblock When optimizing f-divergence is robust with label noise.
\newblock In \emph{International Conference on Learning Representations}, 2021.

\bibitem[Wei et~al.(2021)Wei, Liu, Liu, Niu, and Liu]{wei2021understanding}
Wei, J., Liu, H., Liu, T., Niu, G., and Liu, Y.
\newblock Understanding (generalized) label smoothing when learning with noisy
  labels.
\newblock 2021.

\bibitem[Xia et~al.(2019)Xia, Liu, Wang, Han, Gong, Niu, and
  Sugiyama]{xia2019anchor}
Xia, X., Liu, T., Wang, N., Han, B., Gong, C., Niu, G., and Sugiyama, M.
\newblock Are anchor points really indispensable in label-noise learning?
\newblock In \emph{Advances in Neural Information Processing Systems}, pp.\
  6838--6849, 2019.

\bibitem[Xia et~al.(2020{\natexlab{a}})Xia, Liu, Han, Wang, Deng, Li, and
  Mao]{xia2020extended}
Xia, X., Liu, T., Han, B., Wang, N., Deng, J., Li, J., and Mao, Y.
\newblock Extended {T}: Learning with mixed closed-set and open-set noisy
  labels.
\newblock \emph{arXiv preprint arXiv:2012.00932}, 2020{\natexlab{a}}.

\bibitem[Xia et~al.(2020{\natexlab{b}})Xia, Liu, Han, Wang, Gong, Liu, Niu,
  Tao, and Sugiyama]{xia2020parts}
Xia, X., Liu, T., Han, B., Wang, N., Gong, M., Liu, H., Niu, G., Tao, D., and
  Sugiyama, M.
\newblock Part-dependent label noise: Towards instance-dependent label noise.
\newblock In \emph{Advances in Neural Information Processing Systems},
  volume~33, pp.\  7597--7610, 2020{\natexlab{b}}.

\bibitem[Xia et~al.(2021)Xia, Liu, Han, Gong, Wang, Ge, and
  Chang]{xia2021robust}
Xia, X., Liu, T., Han, B., Gong, C., Wang, N., Ge, Z., and Chang, Y.
\newblock Robust early-learning: Hindering the memorization of noisy labels.
\newblock In \emph{International Conference on Learning Representations}, 2021.

\bibitem[Xiao et~al.(2015)Xiao, Xia, Yang, Huang, and Wang]{xiao2015learning}
Xiao, T., Xia, T., Yang, Y., Huang, C., and Wang, X.
\newblock Learning from massive noisy labeled data for image classification.
\newblock In \emph{Proceedings of the IEEE Conference on Computer Vision and
  Pattern Recognition}, pp.\  2691--2699, 2015.

\bibitem[Xu et~al.(2019)Xu, Cao, Kong, and Wang]{xu2019l_dmi}
Xu, Y., Cao, P., Kong, Y., and Wang, Y.
\newblock L\_dmi: A novel information-theoretic loss function for training deep
  nets robust to label noise.
\newblock In \emph{Advances in Neural Information Processing Systems},
  volume~32, 2019.

\bibitem[Yao et~al.(2020{\natexlab{a}})Yao, Yang, Han, Niu, and
  Kwok]{yao2020searching}
Yao, Q., Yang, H., Han, B., Niu, G., and Kwok, J.~T.
\newblock Searching to exploit memorization effect in learning with noisy
  labels.
\newblock In \emph{Proceedings of the 37th International Conference on Machine
  Learning}, ICML '20, 2020{\natexlab{a}}.

\bibitem[Yao et~al.(2020{\natexlab{b}})Yao, Liu, Han, Gong, Deng, Niu, and
  Sugiyama]{yao2020dual}
Yao, Y., Liu, T., Han, B., Gong, M., Deng, J., Niu, G., and Sugiyama, M.
\newblock Dual t: Reducing estimation error for transition matrix in
  label-noise learning.
\newblock In \emph{Advances in Neural Information Processing Systems},
  volume~33, pp.\  7260--7271, 2020{\natexlab{b}}.

\bibitem[Yu et~al.(2019)Yu, Han, Yao, Niu, Tsang, and Sugiyama]{yu2019does}
Yu, X., Han, B., Yao, J., Niu, G., Tsang, I., and Sugiyama, M.
\newblock How does disagreement help generalization against label corruption?
\newblock In \emph{Proceedings of the 36th International Conference on Machine
  Learning}, volume~97, pp.\  7164--7173. PMLR, 09--15 Jun 2019.

\bibitem[Zhang \& Sabuncu(2018)Zhang and Sabuncu]{zhang2018generalized}
Zhang, Z. and Sabuncu, M.
\newblock Generalized cross entropy loss for training deep neural networks with
  noisy labels.
\newblock In \emph{Advances in neural information processing systems}, pp.\
  8778--8788, 2018.

\bibitem[Zhu et~al.(2021)Zhu, Liu, and Liu]{zhu2020second}
Zhu, Z., Liu, T., and Liu, Y.
\newblock A second-order approach to learning with instance-dependent label
  noise.
\newblock In \emph{The IEEE Conference on Computer Vision and Pattern
  Recognition (CVPR)}, June 2021.

\end{thebibliography}
